\documentclass[final,12pt, times]{elsarticle}
\usepackage{comment}
\usepackage{graphicx}
\usepackage{kotex}
\usepackage{microtype}
\usepackage{lipsum,capt-of,graphicx}% h
\usepackage[margin=1in]{geometry}% Just for this example
\usepackage{subfigure}
\usepackage{placeins}
\usepackage{booktabs} % for professional tables
\usepackage{hyperref}
\usepackage{multirow}
\usepackage{rotating}
\usepackage{makecell}
\usepackage{pifont}
\usepackage{amssymb}
\usepackage{amsmath}
\usepackage[dvipsnames]{xcolor}
\usepackage{enumitem}
\usepackage{bm}
\usepackage{algorithm, algorithmic}
\usepackage[framemethod=tikz]{mdframed}
\usepackage{todonotes}
\usepackage{collcell}
\usepackage{tikz}
\usepackage[acronym]{glossaries}
\usepackage{caption}

\newcommand{\tab}[1]{\hspace{.06\textwidth}\rlap{#1}}

\usepackage{etoolbox}
\usepackage{colortbl,dcolumn}

\newtoggle{inTableHeader}% Track if still in header of table
\toggletrue{inTableHeader}% Set initial value

\newcommand*{\MinNumber}{0.0}%
\newcommand*{\MidNumber}{50.0} %
\newcommand*{\MaxNumber}{100.0}%

\newcommand{\ApplyGradient}[1]{%
  \iftoggle{inTableHeader}{#1}{
    \ifdim #1 pt > \MidNumber pt
        \pgfmathsetmacro{\PercentColor}{max(min(100.0*(#1 - \MidNumber)/(\MaxNumber-\MidNumber),100.0),0.00)} %
        \hspace{-0.33em}\colorbox{green!\PercentColor!yellow}{#1}
    \else
        \pgfmathsetmacro{\PercentColor}{max(min(100.0*(\MidNumber - #1)/(\MidNumber-\MinNumber),100.0),0.00)} %
        \hspace{-0.33em}\colorbox{red!\PercentColor!yellow}{#1}
    \fi
  }}

\def\cca#1{\cellcolor{black!#10}\ifnum #1>5\color{white}\fi{#1}}
\def\cca#1{\cellcolor{black!#10}\ifnum #1>5\color{white}\fi{#1}}
\newcolumntype{R}{>{\collectcell\ApplyGradient}c<{\endcollectcell}}
\renewcommand{\arraystretch}{0}
\journal{Pattern Recognition 115. \href{https://doi.org/10.1016/j.patcog.2021.107899}{\texttt{doi:10.1016/j.patcog.2021.107899}}}

\usepackage{lineno}
\usepackage[normalem]{ulem}
\newcommand{\edited}[2]{{\color{red}\sout{#1}}{\color{blue}{#2}}}   % "full" change info
\renewcommand{\edited}[2]{{#2}}                        % only shows and highlights new parts
\newcommand{\editedtwo}[2]{{\color{blue}{#2}}}
\renewcommand{\editedtwo}[2]{{#2}}

\begin{document}

%NOTE! in order to update the glossary below, one has to force compilation errors once after editing he entries. just uncommend below line, compile, and comment it again
%\newglossaryentry

\glsdisablehyper
\newacronym{cnn}{CNN}{convolutional neural network}
%\newacronym{cnn}{DNN}{Deep Neural Network}
\newacronym{lrp}{LRP}{Layer-wise Relevance Propagation}
\newacronym{dtd}{DTD}{Deep Taylor Decomposition}
\newacronym{flop}{FLOPs}{floating-point operations}
\newacronym{ssl}{SSL}{Structured Sparsity Learning}
\newacronym{mlp}{MLP}{Multi-Layer Perceptron}
\newacronym{lstm}{LSTM}{Long Short-Term Memory Network}
\newacronym{gmacs}{GMACS}{Giga Multiply-Accumulate Operations per Second}
\newacronym{mmacs}{MMACS}{Mega Multiply-Accumulate Operations per Second}

\begin{frontmatter}


\title{Pruning by Explaining: A Novel Criterion for\\ Deep Neural Network Pruning}

\author[label1,label5]{Seul-Ki Yeom}
\ead{yeom@tu-berlin.de}
\author[label1,labelP]{Philipp Seegerer}
\ead{philipp.seegerer@tu-berlin.de}
\author[label2]{Sebastian Lapuschkin}
\ead{sebastian.lapuschkin@hhi.fraunhofer.de}
\author[labelA,labelB]{Alexander Binder}
\ead{alexabin@uio.no}
\author[label2]{\mbox{Simon Wiedemann}}
\ead{simon.wiedemann@hhi.fraunhofer.de}
\author[label1,label3,label4,labelBIFOLD]{Klaus-Robert M{\"u}ller\corref{cor1}}
\ead{klaus-robert.mueller@tu-berlin.de}
\author[label2,labelBIFOLD]{Wojciech Samek\corref{cor1}}
\ead{wojciech.samek@hhi.fraunhofer.de}
\address[label1]{Machine Learning Group, Technische Universit{\"a}t Berlin, 10587 Berlin, Germany}
\address[labelBIFOLD]{BIFOLD -- Berlin Institute for the Foundations of Learning and Data, Berlin, Germany}
\address[label2]{Department of Artificial Intelligence, Fraunhofer Heinrich Hertz Institute, 10587 Berlin, Germany}
\address[labelA]{ISTD Pillar, Singapore University of Technology and Design, Singapore 487372, Singapore}
\address[labelB]{Department of Informatics, University of Oslo, 0373 Oslo, Norway}
\address[label3]{Department of Artificial Intelligence, Korea University, Seoul 136-713, Korea}
\address[label4]{Max Planck Institut für Informatik, 66123 Saarbr{\"u}cken, Germany}
\address[labelP]{Aignostics GmbH, 10557 Berlin, Germany}
\address[label5]{Nota AI GmbH, 10117 Berlin, Germany}
\cortext[cor1]{Corresponding Authors}

\begin{abstract}
% Motivation
The success of \glspl{cnn} in various applications is accompanied by a significant increase in computation and parameter storage costs. Recent efforts to reduce these overheads involve pruning and compressing the weights of various layers while at the same time aiming to not sacrifice performance.
% Contribution of this paper
In this paper, we propose a novel criterion for \gls{cnn} pruning inspired by neural network interpretability: The most relevant units, i.e. weights or filters, are automatically found using their relevance scores obtained from concepts of explainable AI (XAI). By exploring this idea, we connect the lines of interpretability and model compression research.
% Results
We show that our proposed method can efficiently prune \gls{cnn} models in transfer-learning setups in which networks pre-trained on large corpora are adapted to specialized tasks.
The method is evaluated on a broad range of computer vision datasets.
Notably, our novel criterion is not only competitive or better compared to state-of-the-art pruning criteria when successive retraining is performed, but clearly outperforms these previous criteria in the resource-constrained application scenario in which the data of the task to be transferred to is very scarce and one chooses to refrain from fine-tuning.
% Discussion
Our method is able to compress the model iteratively while maintaining or even improving accuracy.
At the same time, it has a computational cost in the order of gradient computation and is comparatively simple to apply without the need for \mbox{tuning hyperparameters for pruning.}
\end{abstract}


\end{frontmatter}

%\linenumbers

\section{Introduction}
\label{Introduction}
%Deep Convolutional Neural Networks (CNNs) are getting deeper and wider to improve their performance and thus increase their computational complexity
    Deep \glspl{cnn} have become an indispensable tool for a wide range of applications~\cite{GuWKMSSLWWCC18}, such as image classification, speech recognition, natural language processing, chemistry, neuroscience, medicine and even are applied for playing games such as Go, poker or \edited{s}{S}uper \edited{s}{S}mash \edited{b}{B}ros. They have achieved high predictive performance, at times even outperforming humans. Furthermore, in specialized  domains where limited training data is available, e.g., due to the cost and difficulty of data generation (medical imaging from fMRI, EEG, PET etc.), transfer learning can improve the CNN performance by extracting the knowledge from the source tasks and applying it to a target task which has limited training data.

	However, the high predictive performance of \glspl{cnn} often comes at the expense of high storage and computational costs, which are \edited{directly}{} related to \edited{}{the} energy expenditure \edited{demanded by the elaborate architecture}{} of the fine-tuned network. These deep architectures are composed of millions of parameters to be trained, leading to overparameterization (i.e. having more parameters than training samples) of the model~\cite{Denil2013Predicting}. The run-times are typically dominated by the evaluation of convolutional layers, while dense layers are cheap but memory-heavy~\cite{Vivienne2017}. For instance, the VGG-16 model has approximately 138 million parameters, taking up more than 500MB in storage space, and needs 15.5 billion \gls{flop} to classify a single image.
	ResNet50 has approx. 23 \editedtwo{thousand}{million} parameters and needs 4.1 billion \gls{flop}. Note that overparametrization is helpful for an efficient and successful training of neural networks, however, once the trained and well generalizing network structure is established, pruning can help to reduce redundancy while still maintaining good performance~\cite{NIPS1989_250}.

	% Why do we need pruning? And what is the important for pruning?
	%Given that neural networks have become larger and larger with more layers and nodes, energy consumption of these models becomes an important challenge, thus
	Reducing a model's storage requirements and computational cost becomes critical for a broader applicability, e.g., in embedded systems, autonomous agents, mobile devices, or edge devices~\cite{TuL19}.
	%
	%Moreover, especially when used with resource-constrained devices such as embedded sensors, autonomous agents, mobile devices, or single board computers where computational and parameter storage costs are limited and energy efficiency becomes more critical.
	Neural network pruning has a decades long history with interest \edited{in}{from} both academia and industry~\cite{cheng2017survey} aiming to \edited{detect and}{} eliminate the subset of \editedtwo{the}{} network \editedtwo{elements}{units} (i.e.~weights or filters) \edited{that include connections and neurons less}{which is the least} important w.r.t.\edited{}{~}the network's intended task. For network pruning, it is \edited{very}{} crucial to decide how to \edited{quantify}{identify} the ``irrelevant'' subset of the parameters \edited{in the current state}{meant} for deletion. To address this issue, previous researches have proposed specific criteria based on \edited{for instance} Taylor expansion, weight, gradient, \edited{etc.}{and others,} to reduce complexity and computation\edited{s}{} costs in the network\edited{and related}{. Related}  works are introduced in Section~\ref{related_work}.
	% However, the survey of~\cite{cheng2017survey} showed that existing criteria require additional $l_p$ regularization for iterative pruning and also needs the setup of sensitivity per layer type and could be cumbersome for some applications.

	From \edited{the}{a} practical point of view, the full capacity (in terms of weights and filters) of an overparameterized model may not be required, e.g., when
	%\begin{enumerate}
	%    \item
	(1) parts of the model lie dormant after training (i.e., are permanently "switched off"),
	%    \item
	(2) a user is not interested in the model’s full array of possible outputs, which is a common scenario in transfer learning (e.g. the user only has use for 2 out of 10 available network outputs), or
    %	  \item
    (3) a user lacks data and resources for fine-tuning and running the overparameterized model.
	%\end{enumerate}

	In these scenarios the redundant parts of the model will still occupy space \edited{on disk and in system}{in} memory, and \edited{simultaneously}{} information will be propagated \edited{throughout the redundant parts of the model}{through those parts}, consuming energy and increasing runtime.
	%This results in an increased runtime and energy consumption for inference, oftentimes far beyond the optimal minimum.
	Thus, criteria able to stably and significantly reduce the computational complexity of deep neural networks \edited{broadly}{} across applications \edited{would be highly versatile}{are relevant} for \edited{all}{} practitioners.

	% Why did we choose \gls{lrp} as criterion? Assigns importance to every element and is directly linked to network output.
	In this paper, we propose a novel pruning framework based on \gls{lrp}~\cite{bach2015pixel}. %note, however, that other explaining methods could also used with straight forward adaptation. % which can be effectively used for .
	 \gls{lrp} was originally developed as an explanation method to assign importance scores, so called {\it relevance}, to the different input dimensions of a neural network that reflect the contribution of an input dimension to the models decision,
	 and has been applied to different fields of computer vision (e.g., \cite{lapuschkin2019unmasking,hgele2019resolving,seegerer2020interpretable}).
	The relevance is backpropagated from the output to the input and hereby assigned to each \editedtwo{element}{unit} of the deep model.
	Since relevance scores are computed for every layer and neuron from the model output to the input, these relevance scores essentially reflect the importance of every single \editedtwo{element}{unit} of a model and its contribution to the information flow through the network --- a natural candidate to be used as pruning criterion.
	The \gls{lrp} criterion %is not merely based on heuristics
	can be motivated theoretically through the concept of \edited{Deep Taylor Decomposition}{\gls{dtd}} (c.f.~\cite{montavon2017explaining,montavon2018methods,samek2020toward}). %thus offering a systematic approach that is not only based on heuristics. % because it is directly linked to the network prediction.
	Moreover, \gls{lrp} is scalable and easy to apply, and has been implemented in software frameworks such as \edited{}{iNNvestigate}~\cite{alber2019innvestigate}.
	Furthermore, it has linear computational cost \edited{}{in terms of network inference cost}, \edited{like}{similar to} backpropagation.

	%% Experiments and results
	We systematically evaluate the compression efficacy of the \gls{lrp} criterion compared to common pruning criteria \edited{on}{for} two different scenarios.\\[+6px]
	% Scenario 1
	{\bf Scenario 1}: We \edited{focus on pruning of}{prune} pre-trained \glspl{cnn} \edited{including}{followed by} subsequent fine-tuning.
	This is the usual setting in \gls{cnn} pruning and requires a sufficient\edited{ly large}{} amount of data and computational power.\\[+3px]
	{\bf Scenario 2}: \edited{We focus on another scenario, in which a model was pre-trained and}{In this scenario a pretrained model} needs to be transferred to a related problem \edited{}{as well,} but the data available for the new task is too scarce for a proper fine-tuning and/or the time consumption, computational power or energy consumption is constrained. Such transfer learning with restrictions is common in mobile or embedded applications.\\[+3px]
	%In contrast to the first scenario, we directly pruned the convolutional layers of pre-trained model on ILSVRC~2012 without further fine-tuning.
	% Dense layer pruning
	%Additionally, we will demonstrate that our proposed criterion also works for fully-connected layers.\\[+6px] %and is not restricted to the convolutional layers.
	% Results with retraining
	Our experimental results on various benchmark datasets and \edited{two}{four} different popular \gls{cnn} architectures show that the \gls{lrp} criterion for pruning is more scalable and efficient, and leads to better performance than existing criteria regardless of data types and model architectures if retraining is performed (\edited{s}{S}cenario 1).
	% Results without retraining
	Especially, if retraining is prohibited due to external constraints after pruning, the \gls{lrp} criterion clearly outperforms previous criteria on all datasets (Scenario 2). Finally\edited{}{,} we would like to note that our proposed pruning framework is not limited to \gls{lrp} \edited{}{and image data}, but can be also used with other explanation techniques \edited{}{and data types}.

	The rest of this paper is organized as follows: Section~\ref{related_work} summarizes related works for network compression and introduces the typical criteria for network pruning. Section~\ref{methods} describes the framework and details of our approach. The experimental results are illustrated and discussed in Section~\ref{Experiments}\edited{}{, while our approach is discussed in relation to previous studies in Section~\ref{Discussion}}. Section~\ref{conclusion} gives conclusions and an outlook to future work.

\section{Related Work}
\label{related_work}
	% And the fully connected layers of VGG-16 occupy 90\% of the total parameters but only contribute less than 1\% of the overall floating point operations (FLOP)

	\edited{Research efforts have been made in the field of network compression and acceleration. For instance,}{We start the discussion of related research in the field of network compression} with network quantization methods \edited{}{which} have been proposed for storage space compression by decreasing the number of possible and unique values for the parameters~\cite{Wiedemann2019TNNLS, TungMori18}.
	Tensor decomposition approaches decompose network matrices into several smaller ones to estimate the informative parameters of the deep \glspl{cnn} with low-rank approximation/factorization~\edited{}{\cite{GuoXXX19}}.
	%~\cite{DingCH19, GuoXXX19}.

	More recently, \cite{XuYZL19} also propose a framework of architecture distillation based on layer-wise replacement, called LightweightNet for memory and time saving.
	Algorithms for designing efficient models focus more on acceleration instead of compression by optimizing convolution operations or architectures directly~\edited{}{(e.g.~\cite{zhang2017shufflenet}).}
	%~\cite{zhang2017shufflenet, howard2017mobilenets}.

	Network pruning approaches remove redundant or irrelevant \editedtwo{element}{unit}s --- i.e. nodes, filters, or layers --- from the model which are not critical for performance \cite{cheng2017survey, MolchanovMTFK19}.
	Network pruning is robust to various settings and \edited{easily and cheaply}{} gives reasonable compression rates while not (or minimally) hurting the model accuracy. Also it can support both training from scratch and transfer learning from pre-trained models. Early works have shown that network pruning is effective in reducing network complexity and simultaneously addressing over-fitting problems. %Thus for applications which require stable model accuracy, network pruning approaches are being utilized typically.
	Current network pruning techniques make weights or channels sparse by removing non-informative connections and require an appropriate criterion for identifying which \editedtwo{element}{unit}s of the model are not relevant for solving a problem. Thus, it is \edited{very}{} crucial to decide how to quantify the relevance of the parameters (i.e. weights or channels) \edited{from the current sample}{in the current state of} \edited{in stochastic}{the} learning \edited{}{process} for deletion without sacrificing predictive performance. In previous studies, pruning criteria have been proposed based on the magnitude of their 1) weights, 2) gradients, 3) Taylor expansion/derivative, and 4) other criteria, as described in the following section.
	%filter pruning의 장점:: Firstly, it does not introduce sparsity to the original network structure, thus requires no special software or hardware implementations for the resulting models. Secondly, it does not require huge disk storage and runtime memory in inference stage. In this reason, there has been a significant amount of works on reducing the storage and computational cost by model pruning approaches. unstructured pruning generates sparse convolutional kernels which can be hard to accelerate given the lack of efficient sparse libraries, especially for the case of low-sparsity.

	\textbf{Taylor expansion:} Early approaches towards neural network pruning --- optimal brain damage~\cite{NIPS1989_250} and optimal brain surgeon~\cite{hassibi1993second} --- leveraged a second-order Taylor expansion based on the Hessian matrix of the loss function to select parameters for deletion. However, computing the inverse of Hessian is \edited{accompanied by high computational effort}{computationally expensive}. The work of~\cite{MolchanovTKAK16, YuWCQ19} used a first-order Taylor expansion as a criterion to approximate the change of loss in the objective function as an effect of pruning away network \editedtwo{element}{unit}s. \edited{}{We contrast our novel criterion to the computationally more comparable first-order Taylor expansion from~\cite{MolchanovTKAK16}.}

	\textbf{Gradient:} \citet{LIU201984} proposed a hierarchical global pruning strategy by calculating the mean gradient of feature maps in each layer. \edited{}{They adopt a hierarchical global pruning strategy between the layers with similar sensitivity.} \edited{}{\citet{sun2017meprop} proposes a sparsified back-propagation approach for neural network training using the magnitude of the gradient to find essential and non-essential features in \edited{}{\gls{mlp}} and \edited{}{\gls{lstm}} models, which can be used for pruning\edited{ as a criterion}{}.
	We implement the gradient-based pruning criterion after~\cite{sun2017meprop}.}

	\textbf{Weight:} A recent trend is to prune redundant, non-informative weights in pre-trained \gls{cnn} models\edited{}{, based on the magnitude of the weights themselves}.
	\edited{}{\citet{NIPS2015_5784}}~and \edited{}{\citet{han2016eie}}~proposed \edited{pruning the weights whose}{the pruning of weights for which the} magnitude is below a certain threshold, and to subsequently fine-tune with a $l_{p}$-norm regularization. This pruning strategy has been used on fully-connected layers and introduced sparse connections with BLAS libraries, supporting specialized hardware to achieve its acceleration.
	In the same context, \gls{ssl} added group sparsity regularization to penalize unimportant parameters by removing some weights~\cite{wen2016learning}.
	\edited{}{\citet{li2016pruning}}\edited{}{, against which we compare in our experiments, }proposed a one-shot channel pruning method using the $l_{p}$ norm of weights for filter selection, provided that those channels with smaller weights always produce weaker activations.
	%
	% STUFF BELOW CAN BE REMOVED TO REDUCE THE NUMBER OF REFS BY 6 !!!! IS THIS ESSENTIAL INFO FOR THE PAPER, OR CAN THIS SUMMARIZED AS "MANY APPROACHES IN LITERATURE..."
	%\cite{hu2016network}~introduced a network trimming approach that removes those channels whose output activations contain most zero values.
	%
	% Further Stripped away weight refs. Moving the last two citations to "other criteria" as they essentially are hybrid approaches.
	%
	%Recently, channel pruning alternatively used LASSO regression based channel selection and feature map reconstruction to prune filters~\cite{he2017channel}.  \cite{anwar2017structured}~performed structured pruning in convolutional layers by considering strided sparsity of feature maps and kernels to avoid the need for custom hardware and uses particle filters to decide the importance of connections and paths.
	%In contrast to previous pruning studies for deep deterministic models, \cite{ZhangZCL19}~proposed a pruning approach for deep probabilistic models by using the mask of the weights.
	%
	% moved below refs to "other criteria"
	% \cite{GAN2020190}~proposed a fusion approach to combine with weight-based channel pruning and network quantization.
	% % \cite{ChangS19}~proposed a weight-based network pruning approach using the modified $l_{1/2}$ penalty to increase the sparsity of the pre-trained models.
	% More recently, \cite{DaiYJ19} proposed evolutionary paradigm for weight-based pruning and gradient-based growing to reduce the network heuristically.

	\textbf{Other criteria:}
	\cite{yu2017nisp} proposed the Neuron Importance Score Propagation (NISP) algorithm to propagate the importance scores of final responses before the softmax, classification layer in the network.
	The method is based on --- in contrast to our proposed metric --- a \edited{}{per-}layer\edited{-independent}{} pruning process which does not consider global importance in the network. \edited{}{\citet{LuoZZXWL19}} proposed \edited{thinet}{ThiNet,}\edited{that is}{} a data-driven statistical channel pruning technique based on the statistics computed from the next layer.
	Further hybrid approaches can be found in, e.g.~\cite{GAN2020190}, which suggests a fusion approach to combine with weight-based channel pruning and network quantization. More recently, \citet{DaiYJ19} proposed \edited{}{an} evolutionary paradigm for weight-based pruning and gradient-based growing to reduce the network heuristically.

\section{LRP-Based Network Pruning}
\label{methods}
	A feedforward \gls{cnn} consists of neurons established in a sequence of multiple layers\editedtwo{ of neurons}{},
	where each neuron receives the input data from \edited{the}{one or more} previous \edited{layer}{layers} and propagates its output to every neuron in the \edited{next}{succeeding} \edited{layer}{layers},
	using a \edited{}{potentially} non-linear mapping.
	Network pruning aims to sparsify \edited{the}{these} \editedtwo{element}{unit}s by eliminating weights or filters that are non-informative (according to a certain criterion). We specifically focus our experiments on transfer learning, where the parameters of a network pre-trained on a \textit{source} domain is subsequently fine-tuned on a \textit{target} domain, i.e., the final data or prediction task.
	Here, the \edited{}{general} pruning procedure \edited{can be described as follows:}{is outlined in Algorithm~\ref{alg:pruning_general}.}

    %OLD FRAMED REPRESENTATION
    \edited{
%	\begin{mdframed}[
%		roundcorner=4pt,
%		]
%		%\begin{pseudocode}[<ovalbox>]{}{}
%		\textbf{\edited{}{Neural} Network Pruning}:
%		\begin{enumerate}
%			\item[\textit{1.}] Given a pre-trained model in the target domain
%			\item[\textit{2.}] Define a pruning criterion
%			\item[\textit{3.}] Repeatedly prune the network as follows:
%			\begin{enumerate}
%				\item[i.] For each layer,
%				\begin{enumerate}
%					\item[a.] For each element (weight/filter), evaluate the importance according to the pruning criterion (compute magnitudes)
%					\item[\textit{b.}] \textit{optional:} Globally scale the magnitudes with regularization (e.g. $l_p$-norm)
%				\end{enumerate}
%				\item[ii.] Sort the magnitudes for all the layers throughout the network
%				\item[iii.] Prune the least important elements and their inputs and outputs
%				\item[\textit{iv.}] \textit{optional:} Further fine-tune to compensate performance degradation
%			\end{enumerate}
%
%			\item[\textit{4.}] Stop pruning if the model is reduced to a desired amount of model size or performance
%		\end{enumerate}
%		%\end{pseudocode}
%	\end{mdframed}
	%END OLD FRAMED REPRESENTATION
	}{}
	%NEW ALG REPRESENTATION
	%\renewcommand{\algorithmiccomment}[1]{//#1} % comments are bullshit and do not work
	\begin{algorithm}[h!]
		\caption{Neural Network Pruning}
		\label{alg:pruning_general}
		\begin{algorithmic}[1]
			\STATE {\bfseries Input:} pre-trained model \texttt{net}, reference data $\textbf{x}_r$, training data $\textbf{x}_t$
			\STATE \tab\tab\ pruning threshold $t$, pruning criterion $c$,  pruning ratio $r$
			\WHILE{$t$ not reached}
    			\STATE{// \textit{Step 1:} assess network substructure importance}
			    \FORALL{\texttt{layer} {\bfseries in} \texttt{net}}
			        \FORALL{\texttt{\editedtwo{element}{units}} {\bfseries in} \texttt{layer}}
			            \STATE $\rhd$ compute \texttt{importance} of \texttt{\editedtwo{element}{unit}} w.r.t. $c$ (and $\textbf{x}_r$)
			        \ENDFOR
			        \IF{required for $c$}
			            \STATE $\rhd$ globally regularize \texttt{importance} per \texttt{\editedtwo{element}{unit}}
			        \ENDIF
			    \ENDFOR
        		\STATE{// \textit{Step 2:} identify and remove least important \editedtwo{elements}{units} in groups of $r$}
                \STATE $\rhd$ remove $r$ \texttt{\editedtwo{amount of element}{units}} from \texttt{net} {\bfseries where} \texttt{importance} is minimal
                \STATE $\rhd$ remove orphaned connections of each removed \texttt{\editedtwo{element}{unit}}
                \IF{desired}
                    \STATE{// \textit{Step 2.1:} optional fine-tuning to recover performance}
                    \STATE{ $\rhd$ fine-tune \texttt{net} on $\textbf{x}_t$}
                \ENDIF
			\ENDWHILE
			\STATE{// return the pruned network upon hitting threshold $t$ (e.g. model performance or size)}
			\STATE {\bfseries return} \texttt{net}
		\end{algorithmic}
	\end{algorithm}
	%END NEW ALG REPRESENTATION

	%
	%\begin{enumerate}[label=\arabic*)]
	%	\item Given a pre-trained network in the target domain,
	%	\item Prune according to the criterion rule,
	%	\begin{enumerate}
	%		\item Evaluate the magnitude (a.k.a. importance) of each weight/filter layer-wisely by criterion
	%		\item Globally scale the magnitudes with regularization (e.g. $l_p$-norm)
	%		\item Sort the magnitudes for all the layers throughout network
	%		\item Prune the less important nodes/filters and its connections from the lower/upper layers
	%	\end{enumerate}
	%	\item Further fine-tune to compensate performance degradation,
	%	\item Iteratively perform 2) and 3),
	%	\item Stop pruning until the desired amount of pruning is reached based on the proper model size, performance, etc.
	%\end{enumerate}
	Even though most approaches use an identical process, choosing a suitable pruning criterion to quantify the importance of model parameters for deletion while minimizing performance drop~(\edited{Step 3}{Step 1}) is of critical importance, governing the success of the approach.

	\subsection{Layer-wise Relevance Propagation}
	\label{sec:proposed_method}
	%Introduction of LRP
	In this paper, we propose a novel criterion for pruning neural network \editedtwo{element}{unit}s: the \emph{relevance} quantity computed with \gls{lrp}~\cite{bach2015pixel}. \gls{lrp} decomposes a classification decision into \edited{}{proportionate} contributions \edited{called ``relevances''}{}of each network \editedtwo{element}{unit} to the overall classification score\edited{}{, called ``relevances''}.
	% What has \gls{lrp} originally been used for? As interpretability method
    When computed for the input dimensions of a \gls{cnn} and visualized as a heatmap, these relevances highlight parts of the input that are important for the classification decision.
    \gls{lrp} thus originally served as a tool \edited{ to}{for} interpret\edited{}{ing} non-linear learning machines and has been applied as such in various fields, amongst others for general image recognition, medical imaging and natural language processing\edited{~}{, cf.~\cite{samek2019explainable}.}
    %~\cite{lapuschkin2019unmasking}.
    % Why do we want to use it for pruning?
    \edited{However, t}{T}he direct linkage of the relevances to the classifier output\edited{}{, as well as the conservativity constraint imposed on the propagation of relevance between layers,} makes \gls{lrp} not only attractive for model explaining, but can also naturally serve as pruning criterion (see \edited{s}{S}ection~\ref{sec:experiments/toy_example}). %; see section~\ref{sec:experiments/toy_example} for a toy example why \gls{lrp} is an intuitive criterion for pruning.

    % More detail on \gls{lrp}
    The main characteristic of \gls{lrp} is a backward pass through the network during which the network output is redistributed to all \editedtwo{element}{unit}s of the network in a layer-by-layer fashion.
    This backward pass is structurally similar to gradient backpropagation and has therefore a similar runtime.
    The redistribution is based on a \textit{conservation principle} such that the relevances can immediately be interpreted as the contribution that a \editedtwo{element}{unit} makes to the network output, hence establishing a direct connection to the network output and thus its predictive performance.
    Therefore, as a pruning criterion, the method is efficient\edited{(similar runtime as gradient backpropagation}{} and easily scalable to generic network structures.
    Independent of the type of neural network layer --- that is pooling, fully-connected, convolutional layers --- \gls{lrp} allows to quantify the importance of \editedtwo{element}{unit}s throughout the network\edited{~\cite{li2016pruning}}{, given a global prediction context}.
    % TODO this paragraph is confusing. Why cite li2016pruning?

%    Main characteristic of relevance as a metric is the conservation property, where each neuron receives a share of the network output, and redistributes it to its predecessors in equal amount until the input variables are reached. Therefore, as a pruning criteron, our method is efficient and easy to scalable to general network structures including kernels, as well as complex DNN models. This principle ensures that the network output activation is fully redistributed through the layers of the network toward the input layer. Independent from the type of neural network layer -- that is pooling, fully-connected, convolutional layers which mostly denominate the computational costs in CNNs -- we can easily select some redundant nodes/filters throughout the network by adopting LRP as a criterion without further regularization.

    \begin{figure}[t]
		\vskip 0.2in
		\begin{center}
			\centerline{\includegraphics[width=.85\columnwidth]{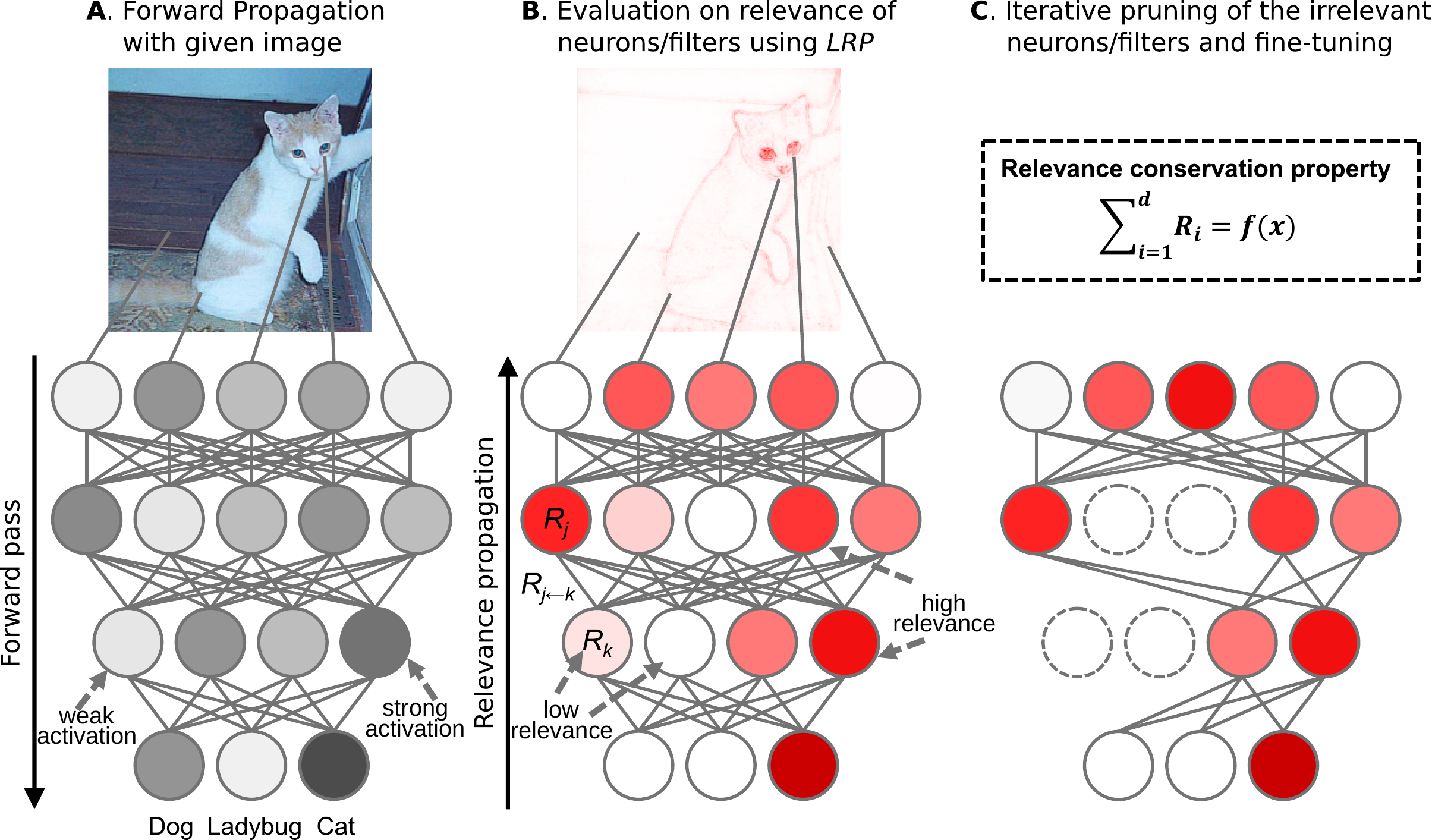}}
			\caption{Illustration of \gls{lrp}-based sequential process for pruning. \textbf{A.} Forward propagation of a given image (i.e. cat) through a  pre-trained model. \textbf{B.} Evaluation on relevance for weights/filters using \gls{lrp}, \textbf{C.} Iterative pruning by eliminating the least relevant \editedtwo{elements}{units} (depicted by circles) and fine-tuning if necessary. The \editedtwo{elements}{units} can be individual neurons, filters, or other arbitrary grouping of parameters, depending on the model architecture.}
			\label{flow_of_LRP}
		\end{center}
		\vskip -0.2in
	\end{figure}

	\subsection{\gls{lrp}-based Pruning}
	% Flow of LRP processing
	The procedure of \gls{lrp}-based pruning is summarized in Figure~\ref{flow_of_LRP}.
	In the first phase, a standard forward pass is performed by the network and the activations at each layer are collected.
    In the second phase, the score \edited{}{$f(\textbf{x})$} obtained at the output of the network\edited{$f(\textbf{x})$}{} is propagated backwards through the network according to \gls{lrp} propagation rules~\cite{bach2015pixel}.
    In the third phase, the current model is pruned by eliminating the irrelevant (\edited{}{w.r.t. the ``relevance'' quantity $R$ obtained via LRP}) \editedtwo{neurons/filters}{units} and is \edited{}{(optionally)} further fine-tuned.
    %Here due to the redistribution of relevance based on \gls{lrp} rules, a set of the highly contributed neurons by prediction have increased relevance scores after pruning.

	\gls{lrp} is based on a layer-wise conservation principle that allows the propagated quantity (e.g. relevance for a predicted class) to be preserved between neurons of two adjacent layers. Let $R_i^{(l)}$ be the relevance of neuron $i$ at layer $l$ and $R_j^{(l+1)}$ be the relevance of neuron $j$ at the next layer $l+1$. Stricter definitions of conservation that involve only subsets of neurons can further impose that relevance is locally redistributed in the lower layers and we define $R_{i \leftarrow j}^{(l)}$ as the share of $R_j^{(l+1)}$ that is redistributed to neuron $i$ in the lower layer. The conservation property always satisfies
	\begin{equation}
    	\label{eq:propery_1}
    	\sum_{i}R_{i \leftarrow j}^{(l)}=R_j^{(l+1)} \enspace ,
	\end{equation}
	where the sum runs over all neurons $i$ of the (during inference) preceeding layer $l$.
	When using relevance as a pruning criterion, this property helps to preserve its quantity layer-by-layer, regardless of hidden layer size
	%(proof: $\sum^{D}\sum_{i}R_{i}=\sum^{D}\sum_{j}R_{j}$ where $D$ denotes the total number of data)
	and the number of iteratively pruned neurons for each layer.
	%(proof: $\sum_{i}^{N_l-n_l}R_{i}=\sum_{j}^{N_{l+1}-n_{l+1}}R_{j} = f(\textbf{x})$ ($n_{l}$: the number of pruned neurons at layer $l$)
	At each layer $l$, we can extract node $i$’s global importance as
	its attributed relevance $R^{(l)}_i$.

% 	Likewise, neurons in the lower layer aggregate all relevances coming from the neurons from the higher layer:

% 	\begin{equation}
% 	\label{eq:propery_2}
% 	R_j=\sum_{k}R_{j \leftarrow k}
% 	\end{equation}
% 	These two equations, \ref{eq:propery_1} and \ref{eq:propery_2}, when combined, also ensure a relevance conservation property between layers (proof: $\sum_{j}R_{j}=\sum_{j}\sum_{k}R_{j \leftarrow k}=\sum_{k}\sum_{j}R_{j \leftarrow k}=\sum_{k}R_{k}$).

	In this paper, we specifically adopt relevance quantities computed with the \gls{lrp}-$\alpha_1\beta_0$-rule as pruning criterion.
	The \gls{lrp}-$\alpha\beta$-rule was developed with feedforward-DNNs with ReLU activations in mind and assumes positive (pre-softmax) logit activations $f_{\text{\tiny logit}}(\textbf{x}) > 0$ for decomposition.
	The rule has been shown to work well in practice in such a setting~\cite{samek2017eval}.
	\editedtwo{}{This particular variant of \gls{lrp} is tightly rooted in \gls{dtd}~\cite{montavon2017explaining}, and other than the criteria based on network derivatives we compare against~\cite{sun2017meprop,MolchanovTKAK16}, always produces \emph{continuous explanations}, even if backpropagation is performed through the discontinuous (and commonly used) ReLU nonlinearity~\cite{montavon2018methods}.
	When used as a criterion for pruning, its assessment of network unit importance will change less abruptly with (small) changes in the choice of reference samples, compared to gradient-based criteria.}

	The propagation rule performs two separate
	relevance propagation steps per layer:
	one exclusively considering activatory parts of the forward propagated quantities (i.e.\ all \mbox{$a_i^{(l)}w_{ij} > 0$})
	and another only processing the inhibitory parts
	(\mbox{$a_i^{(l)}w_{ij} < 0$}) which are subsequently merged in a sum with components weighted by $\alpha$ and $\beta$ (s.t. $\alpha+\beta=1$) respectively.

	By selecting $\alpha=1$, the propagation rule \edited{expresses as}{simplifies to}
	\begin{equation}
	\label{eq:ab_rule_2}
	R_{i}^{(l)} = \sum_{j}  \frac{\edited{}{\left(a_i^{(l)} w_{ij}\right)^{+}}}{\sum_{i'}\edited{}{\left({a_{i'}}^{(l)} w_{i'j}\right)^{+}}} R_j^{(l+1)} \enspace ,
	\end{equation}
	where $R_{i}^{(l)}$ denotes relevance attributed to \edited{}{the}  $i^{th}$ neuron at layer $l$, as an aggregation of downward-propagated relevance messages $R_{i \leftarrow j}^{(l,l+1)}$.
	The \edited{variables $w_{ij}^{+}$ and $a_i^{(l)}$}{terms $\left( \cdot \right)^{+}$} indicate the positive part of the \edited{weight filter and the (strictly positive)}{forward propagated pre-}activation \edited{of the $i^{th}$ neuron at}{from} layer $l$, \edited{respectively}{to layer $(l+1)$}.
	\edited{}{The $i'$ is a running index over all input activations $a$.}
	Note that a choice of $\alpha=1$ only decomposes w.r.t. the parts of the inference signal \emph{supporting}
	the model decision \emph{for} the class of interest.

	\edited{Note that }{}Equation~\eqref{eq:ab_rule_2} is \textit{locally conservative}, i.e. no quantity of relevance gets lost or injected during the distribution of $R_j$ where each term of the sum corresponds to a relevance message $R_{j \leftarrow k}$. For this reason, \gls{lrp} has the following technical advantages over other pruning techniques such as gradient-based or activation-based methods:
	%\begin{enumerate}
	%\item
	(1) Localized relevance conservation implicitly ensures layer-wise\edited{ly}{} regularized global redistribution of importances from each network \editedtwo{element}{unit}.
	%\item
	(2) By summing relevance \edited{over}{} within each (convolutional) filter channel, the \gls{lrp}-based criterion is directly applicable as a measure of total relevance per node/filter, without requiring a post-hoc layer-wise renormalization, e.g., via $l_p$ norm.
	% TODO Summing over channel for per-node pruning?
	%\item
	(3) The use of relevance scores is not restricted to a global application of pruning but can be easily applied to locally and (neuron- or filter-)group-wise constrained pruning without regularization.
    Different strategies for selecting (sub-)parts of the model might still be considered, e.g., applying different weightings/priorities for pruning different parts of the model: Should the aim of pruning be the reduction of \gls{flop} required during inference, one would prefer to focus on primarily pruning \editedtwo{element}{unit}s of the convolutional layers.
    In case the aim is a reduction of the memory requirement,
    pruning should focus on the fully-connected layers instead.
    %\end{enumerate}

	%The variable $W$ denotes the weight matrix connecting the neurons of the two consecutive layers, and $W_{+}$ is the matrix retaining only the positive weights of $W$ setting remaining weights to zero. This vector form is useful when implementing \gls{lrp} for fully connected layers.
	\edited{
	Algorithm~\ref{alg:example} provides an overview of the \gls{lrp}-based pruning process.
	\begin{algorithm}[b!]
		\caption{\gls{lrp}-based pruning}
		\label{alg:example}
		\begin{algorithmic}[1]
			\STATE {\bfseries Input:} pre-trained model $f$,
			\STATE \tab{data $\textbf{x}$, pruning rate $pr$,}
			\STATE \tab{total number of weights/filters $m$}
			\WHILE{pruning}
			\STATE Initialization of storage, $\textit{S}$
			\STATE $\textit{R}$ = $f(\textbf{x})$ \qquad \qquad \tab{// Forward processing}
			%\CMT{Forward processing}
			\FOR{$layer$ {\bfseries in} $f$$[::-1]$}

			\STATE $\textit{R}$ = LRP($layer$, $\textit{R}$)
			\tab{// \gls{lrp} processing}
			%\CMT{\gls{lrp} processing}
			\IF{$layer$ {\bfseries is} `$conv$'}
			\STATE Compute filter-wise sum of $\textit{R}$
			\ENDIF
			\STATE Stack $\textit{R}$ in $\textit{S}$
			\ENDFOR
			\FOR{{\bfseries all} $j$ {\bfseries where} $\textit{S}_j$ $<$ $m$ $\times$ $pr$}
			\STATE Remove the $j^{th}$ weight/filter and its consecutive connections
			\ENDFOR
			\STATE Fine-tuning (optional)
			\ENDWHILE
			\STATE {\bfseries Output:} pruned model $f^{'}$
		\end{algorithmic}
	\end{algorithm}
	}{In the context of Algorithm~\ref{alg:pruning_general}, Step 1 of the \gls{lrp}-based assessment of neuron and filter importance is performed as a single \gls{lrp} backward pass through the model, with an aggregation of relevance \emph{per filter channel} as described above, for convolutional layers, and does not require additional normalization or regularization.}
	\editedtwo{}{We would like to point out that instead of backpropagating the model output $f_c(x)$ for the true class $c$ of any given sample $x$ (as it is commonly done when \gls{lrp} is used for \emph{explaining} a prediction~\cite{bach2015pixel, lapuschkin2019unmasking}), we initialize the algorithm with $R_c^{(L)} = 1$ at the output layer $L$.
	We thus gain robustness against the model's (in)confidence in its predictions on the previously unseen reference samples $x$ and ensure an equal weighting of the influence of all reference samples in the identification of relevant neural pathways.}

\section{Experiments}
\label{Experiments}
    \edited{In the following, we will first attempt to}{We start by an attempt to} intuitively illuminate
    the properties of different pruning criteria\edited{--- that is;}{, namely,} weight magnitude, Taylor, gradient and \gls{lrp},  \edited{--- at hand of}{via} a series of toy datasets.
	We then show the effectiveness of the \gls{lrp} criterion for pruning on \edited{current and}{} widely-used \edited{I}{i}mage recognition benchmark datasets --- i.e. the Scene 15~\cite{lazebnik2006beyond}, Event 8~\edited{}{\cite{Lievent8}}, Cats \& Dogs~\cite{asirra2007}, Oxford Flower 102~\cite{nilsback2008automated}, C\edited{ifar 10}{IFAR-10\footnote{\edited{}{\url{https://www.cs.toronto.edu/\~kriz/cifar.html}}}},
	%~\edited{}{\cite{krizhevsky2009learning}},
	and ILSVRC~2012~\cite{ILSVRC15} datasets --- and \edited{two}{four} pre-trained feed-forward deep neural network architectures, \edited{the}{} AlexNet and \edited{the}{} VGG-16 \edited{}{with only a single sequence of layers, and \edited{the}{}ResNet-18 and ResNet-50~\cite{he2016deep}, which both contain multiple parallel branches of layers and skip connections}.

	%simulated data as well as two different real-world scenarios.
	% First scenario: common scenario
	The first scenario focuses specifically on pruning of pre-trained \glspl{cnn} with subsequent fine-tuning, as it is common in pruning research~\cite{MolchanovTKAK16}.
	We compare our method with several state-of-the-art criteria to demonstrate the effectiveness of \gls{lrp} as a pruning criterion in \glspl{cnn}.
	% Second scenario: not so common
	In the second scenario, we tested whether the proposed pruning criterion also works well if only a very limited number of samples is available for pruning the model.
% 	In contrast to the first scenario, no subsequent fine-tuning is applied due to the limited number of samples.
	This is relevant in case of devices with limited computational power, energy and storage such as mobile devices or embedded applications.
%%	In contrast to the first scenario, we directly pruned the convolutional layers of pre-trained model on ILSVRC~2012 without further fine-tuning.
%%	This is shown by using 1) the Cats and Dogs dataset to train VGG-16 pre-trained on ILSVRC~2012, and 2) small number of images of a small randomly-selected subset of classes from the ILSVRC~2012 validation set.
	% Fully connected layer pruning. Disentangle from previous experiment because otherwise it's hard to justify why we did not prune FC layers in the first scenario.
%%	Additionally, we demonstrate that our proposed criterion also works for fully-connected layers and is not restricted to the convolutional layers.
	%Note that to regain the initial performance between greedy pruning process, we further fine-tune the pruned model 10 epochs with different random seeds, and compute the average performance.

	\subsection{Pruning Toy Models}
	\label{sec:experiments/toy_example}
% 	Before we start on two main scenario, w
	First\edited{ly}{}, we systematically compare the properties and effectiveness of the different pruning criteria on \edited{}{several} toy datasets
	\edited{}{in order to foster an intuition about the properties of all approaches, in a controllable and computationally inexpensive setting}.
    \edited{In order to}{To this end we} evaluate \edited{the}{all four} criteria \edited{under the}{on} different \edited{}{toy} data distributions \edited{}{qualitatively and quantitatively}.
    \edited{we tested all four pruning criteria on}{We generated three $k$-class toy datasets (``moon'' ($k=2$), ``circle'' ($k=2$) and ``multi\edited{-class}{}'' ($k = 4$)),} \edited{as generated}{} using \edited{the}{}respective \edited{}{generator} functions\footnote{\edited{}{\url{https://scikit-learn.org/stable/datasets}}}$^,$\footnote{\edited{}{\url{https://github.com/seulkiyeom/LRP\_Pruning\_toy\_example}}}\edited{from the scikit-learn machine learning toolkit for python and shown in Figure~\ref{plot:toy_data_experiment}}{}.

    \edited{We}{Each} generated \edited{a}{}2D dataset \edited{which}{}consists of 1000 training samples \edited{and 5 test samples}{}per class. We constructed \edited{}{and trained} the model\edited{}{s} as \edited{follows,
    %\begin{enumerate}
    %    \item Linear (2 $\times$ 1000), ReLU
    %    \item Dropout with 0.5
    %    \item Linear (1000 $\times$ 1000), ReLU
    %    \item Linear (1000 $\times$ 1000), ReLU
    %    \item Linear (1000 $\times$ $k$)
    %\end{enumerate}
    }{a sequence of three consecutive ReLU-activated dense layers with 1000 hidden neurons each. After the first linear layer, we have added a DropOut layer with a dropout probability of 50\%. The model receives inputs from $\mathrm{R}^2$ and has --- depending on the toy problem set --- $k \in \lbrace 2,4 \rbrace$ output neurons:}
    \begin{center}
        \texttt{Dense(1000) -> ReLU -> DropOut(0.5) -> Dense(1000) -> \\
                -> ReLU -> Dense(1000) -> ReLU -> Dense(k)}
    \end{center}

    \edited{}{We then sample a number of new datapoints (unseen during training)
    %as ``test samples''
    %for the computation of the models' generalization performance, but also
    for the computation of the pruning criteria.}
    During pruning, we \edited{pruned}{removed a fixed number of} 1000 of \edited{}{the} 3000 \edited{neurons}{\emph{hidden neurons}} that have the least relevance for prediction according to each criterion.
    \edited{}{This is equivalent to removing 1000 learned (yet insignificant, according to the criterion) filters from the model.}
    After pruning, we observed the \edited{change}{changes} in the decision boundaries and re-evaluated \edited{}{for} classification accuracy using the original training samples \edited{}{and re-sampled datapoints} across criteria.
    \edited{}{This experiment is performed with $n \in [1,\editedtwo{}{2,}5,10,20,50,100,200]$ reference samples for testing and the computation of pruning criteria. Each setting is repeated 50 times, using the same set of random seeds (depending on the repetition index) for each $n$ across all pruning criteria to uphold comparability.}

\begin{figure}[ht!]
		\begin{center}
			\centerline{\includegraphics[width=0.99\linewidth]{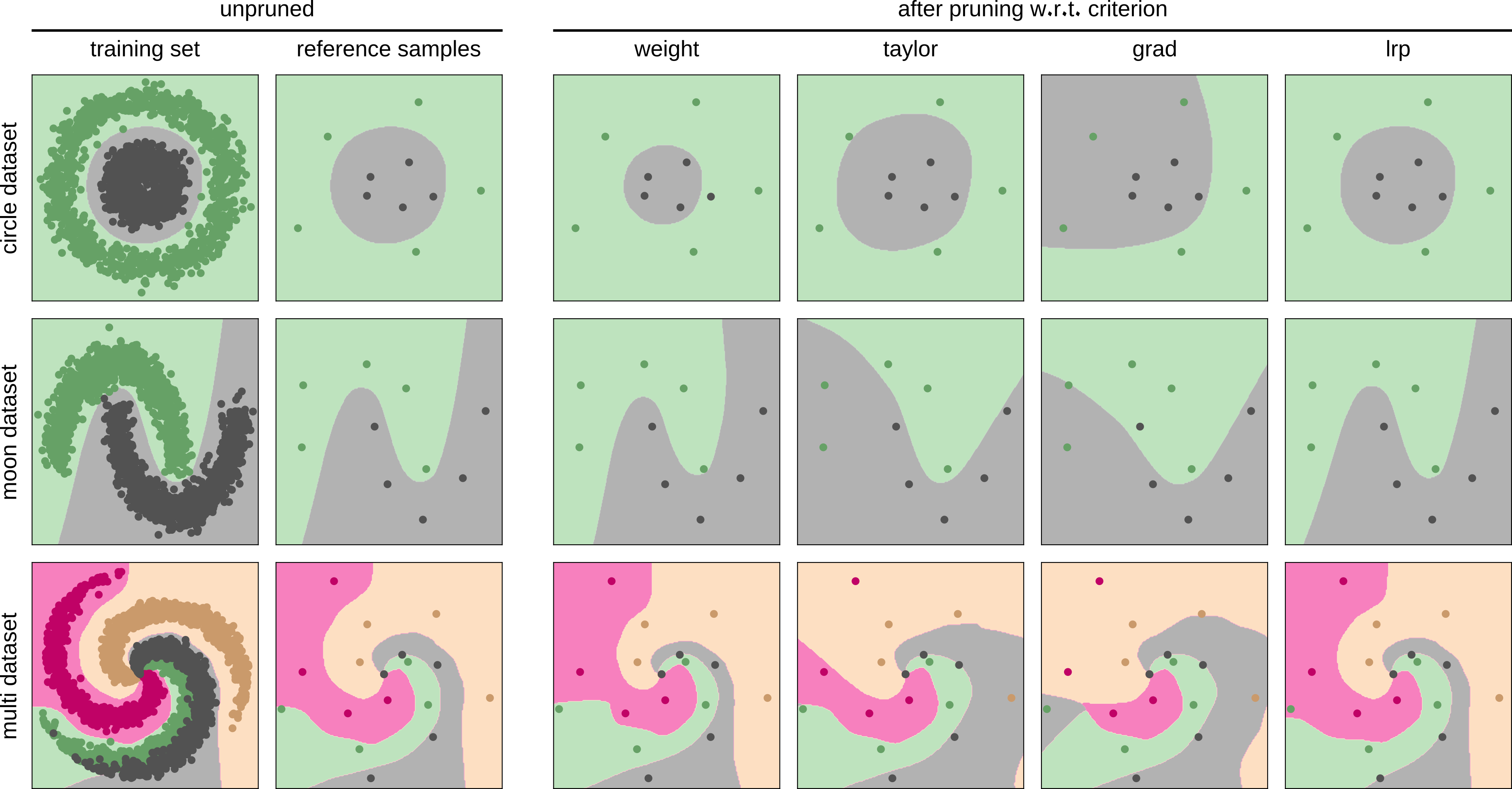}} %feel free to replace *-test with *-train
			\caption{\edited{Performance comparison of the criteria on toy datasets (moon, circle, and multi-class dataset)}{Qualitative comparison of the impact of the pruning criteria on the decision function on three toy datasets.
			%, when pruning w.r.t criteria computed for only 5 reference samples per class.
			}
			\textit{1st column}: scatter plot \edited{}{of the training data} and decision boundary of \edited{}{the} trained model,
			\textit{2nd column}: data samples \edited{}{randomly selected} for \edited{}{computing the} pruning \edited{}{criteria},
			\textit{3rd to 6th columns}: changed decision boundaries \edited{of}{after the application of pruning w.r.t.} different criteria.}
			\label{plot:toy_data_experiment_qualitative}
		\end{center}
	\end{figure}

\begin{figure}[ht!]
    		\begin{center}
    			\centerline{\includegraphics[width=.99\linewidth]{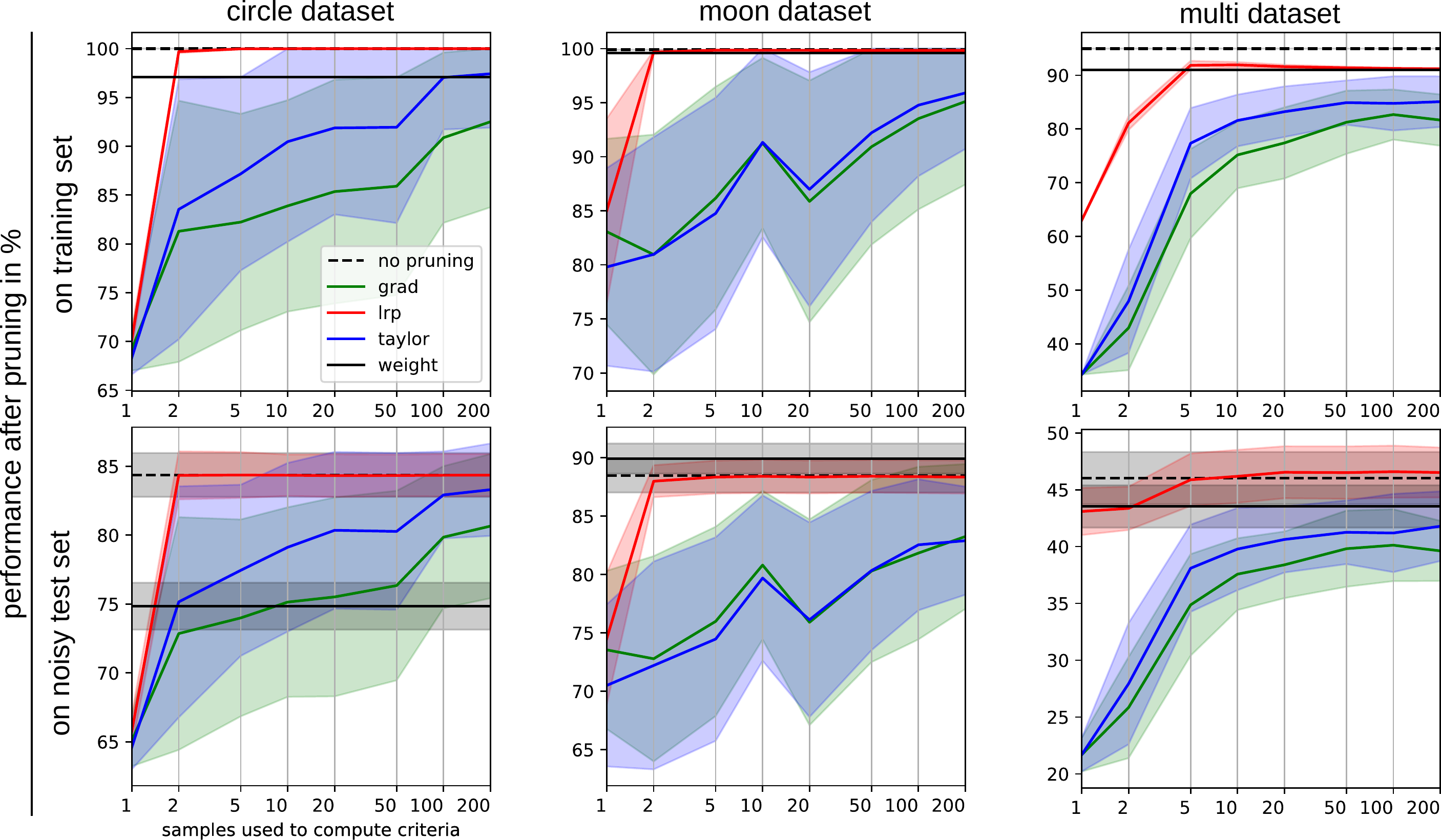}}
    			\caption{\edited{}{Pruning performance (accuracy) comparison of criteria depending on the number of reference samples per class used for criterion computation.
    			\textit{1st row:} Model evaluation on the training data.
    			\textit{2nd row:} Model evaluation on an unseen test dataset with added Gaussian noise ($\mathcal{N}(0,0.3)$), which have not been used for the computation of pruning criteria. Columns: Results over different datasets. Solid lines show the average post-pruning performance of the models pruned w.r.t. to the evaluated criteria weight (black), Taylor (blue), grad(ient) (green) and \gls{lrp} (red) over 50 repetitions of the experiment. The dashed black line indicates the model's evaluation performance without pruning. Shaded areas around the lines show the standard deviation over the repetition of experiments. Further results for noise levels $\mathcal{N}(0,0.1)$ and $\mathcal{N}(0,0.01)$ are available on \href{https://github.com/seulkiyeom/LRP_Pruning_toy_example}{github}$^3$.}
    			}
    			\label{plot:toy_data_experiment_influence_of_n}
    		\end{center}
    	\end{figure}

\begin{figure}[!ht]
        \begin{center}
            \includegraphics[width=\textwidth]{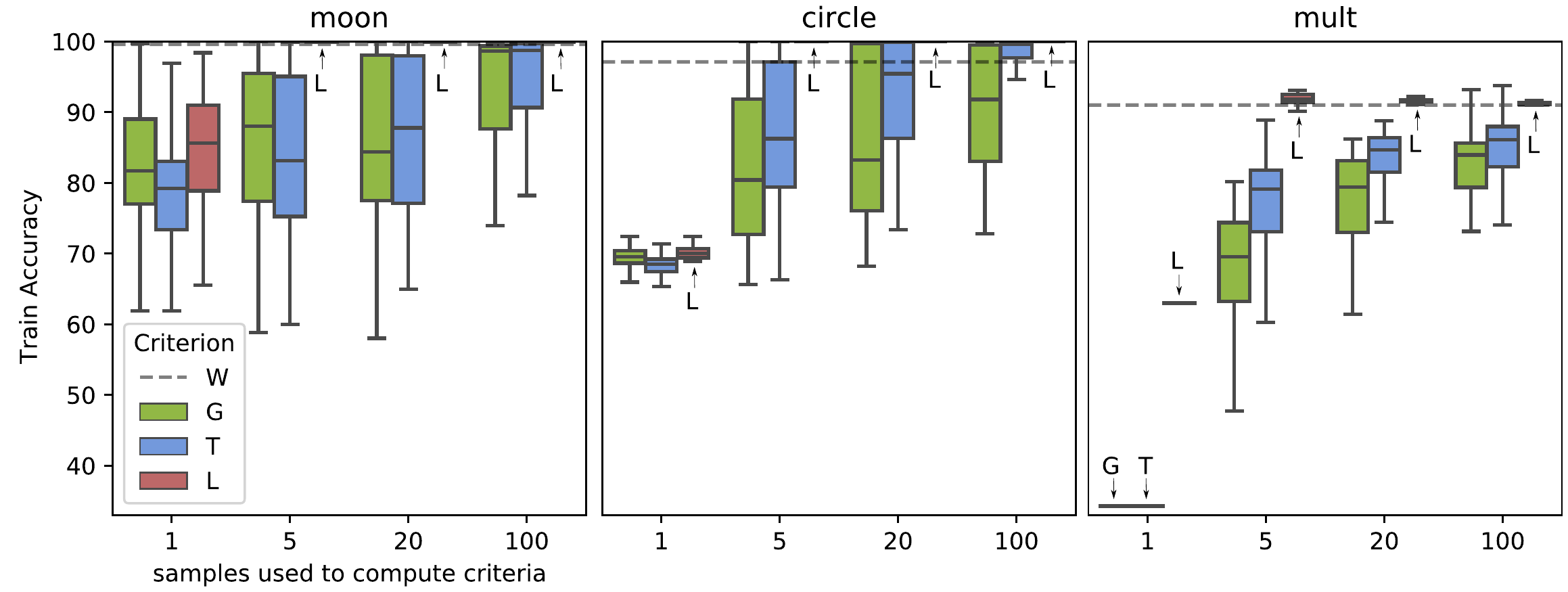}
            \caption{Comparison of \edited{}{training} accuracy \edited{in each criterion with pruned models on toy datasets (moon, circle, and multi-class dataset)}{after one-shot pruning one third of all filters w.r.t one of the four metrics on toy datasets, with $n \in [1, 5, 20, 100]$ reference samples used for criteria computation for \editedtwo{\underline{T}aylor, \underline{G}radient and \underline{L}RP}{\underline{W}eight, \underline{G}radient, \underline{T}aylor and \underline{L}RP}. \editedtwo{The reported accuracies are averages over 50 repetitions in percent.}{The experiment is repeated 50 times.} Note that the \underline{W}eight criterion is
            % static
            %\editedtwo{}{(resulting in  99.60\%, 97.10\%, 91.00\% for the three datasets, respectively, after pruning)}
            %and
            not influenced by the number of reference samples $n$}.
            \editedtwo{}{Compare to Supplementary Table 1.}}
            \label{fig:compareresult_toy}
    		\setlength{\tabcolsep}{4pt}
    		\renewcommand{\arraystretch}{.55}
    		\scriptsize %scriptsize instead of resizebox
    		%\resizebox{\linewidth}{!}{
            %} %closing bracket for \resizebox
        \end{center}
    \end{figure}

    \edited{}{Figure~\ref{plot:toy_data_experiment_qualitative} shows the data distributions of the generated toy datasets, an exemplary set of $n=5$ samples generated for
    %testing and
    criteria computation, as well as the qualitative impact to the models' decision boundary when removing a fixed set of 1000 neurons as selected via the compared criteria.
    Figure~\ref{plot:toy_data_experiment_influence_of_n} investigates how the pruning criteria preserve the models' problem solving capabilities as a function of the number of samples selected for computing the criteria.
    \editedtwo{}{Figure~\ref{fig:compareresult_toy}} then quantitatively summarizes the results for specific numbers of unseen samples ($n \in [1, 5, 20, 100]$) for computing the criteria. Here we report the model accuracy on the \emph{training set} in order to relate the preservation of the \emph{decision function as learned from data} between unpruned (2nd column) to pruned models and pruning criteria (remaining columns).
    }

    \edited{
    On the first toy example (``moon'' dataset, first row in Figure~\ref{plot:toy_data_experiment}), the original model accuracy is 99.9\%. After pruning, the accuracies are 99.6\% (weight), 74.5\% (Taylor), 76.9\% (gradient), and 99.85\% (\gls{lrp}), respectively.
    On the second example (``circle'' dataset, second row), the original accuracy is 100.0\%. After pruning, the accuracies are 97.1\% (weight), 97.25\% (Taylor), 75.05\% (gradient), and 100.0\% (\gls{lrp}), respectively. On the third example (``multi-class'' dataset, third row), the original model accuracy is 94.95\%. After pruning, the accuracies are 91.0\% (weight), 85.15\% (Taylor), 84.93\% (gradient), and 91.3\% (\gls{lrp}), respectively. (See the \edited{ Figure~\ref{fig:compareresult_toy}})
    }
    {
    %Table~\ref{fig:compareresult_toy} summarizes the results on all three toy datasets.
    %That is, the table lists the dataset names in the first column and the original model accuracy in the second column.
    %Obtained accuracy scores \emph{after} the application of each of the four pruning critera are listed in the remaining columns, respectively.
    }

    \edited{}{
    The results in \editedtwo{}{Figure~\ref{fig:compareresult_toy}} show that, among all criteria based on reference sample for the computation of relevance,
    the \gls{lrp}-based measure consistently outperforms all other criteria in all reference set sizes and datasets.
    Only in the case of $n=1$ reference sample per class, the weight criterion preserves the model the best.
    Note that using the weight magnitude as a measure of network \editedtwo{element}{unit} importance is a static approach, independent from the choice of reference samples.
    Given  $n=5$ points of reference per class, the \gls{lrp}-based criterion already outperforms also the weight magnitude as a criterion for pruning unimportant neural network structures,
    while successfully preserving the \emph{functional core} of the predictor.
    Figure~\ref{plot:toy_data_experiment_qualitative} demonstrates how the toy models' decision boundaries change under influence of pruning with all four criteria.
    We can observe that the weight criterion and \gls{lrp} preserve the models' learned decision boundary well. Both the Taylor and gradient measures degrade the model significantly.
    Compared to weight- and \gls{lrp}-based criteria, models pruned by gradient-based criteria misclassify a large part of samples.
    }

    \edited{}{
    The first row of Figure~\ref{plot:toy_data_experiment_influence_of_n} shows that all (data dependent) measures benefit from increasing the number of reference points.
    \gls{lrp} is able to find and preserve the functionally important network components with only very little data, while at the same time being considerably less sensitive to the choice of reference points than other metrics, visible in the measures' standard deviations.
    Both the gradient and Taylor-based measures do not reach the performance of \gls{lrp}-based pruning, even with 200 reference samples for each class.
    The performance of pruning with the weight magnitude based measure is constant, as it does only depend on the learned weights itself.
    }
    %
    %\edited{}{
    %In the case of the pruned models being evaluated on the unseen and noisy test samples (bottom row of Figure~\ref{plot:toy_data_experiment_influence_of_n}), which themselves have been used for the computation of the (data dependent) pruning criteria,
    %we can see the standard deviation across repetitions increase for all measures (as well as the unpruned %model), including LRP, but notably also the criterion based on the weights themselves.
    %This is primarily caused by mispredictions on the newly generated samples and less so through the pruning.
    %In this case the LRP-based criterion still outperforms the other measures, or at least is in line with the weight-based pruning, while demonstrating an overall lower standard deviation.
    %}
    \edited{}{
    The bottom row of Figure~\ref{plot:toy_data_experiment_influence_of_n} shows the test performance of the pruned models as a function of the number of samples used for criteria computation.
    Here, we tested on 500 samples per class, drawn from the datasets' respective distributions, and perturbed with additional gaussian noise ($\mathcal{N}(0,0.3)$) added after data generation.
    Due to the large amounts of noise added to the data, we see the prediction performance of the pruned and unpruned models to decrease in all settings.
    Here we can observe that two out of three times the \gls{lrp}-pruned models outperforming all other criteria. Only once, on the ``moon'' dataset, pruning based on the weight criterion yields a higher performance than the \gls{lrp}-pruned model.
    Most remarkably though, only the models pruned with the \gls{lrp}-based criterion exhibit prediction performance and behavior --- measured in mean and standard deviation of accuracies measured over all 50 random seeds per $n$ reference samples on the deliberatly heavily noisy data --- highly similar to the original and unpruned model, from only $n=5$ reference samples per class on, \editedtwo{}{on all datasets}.
    This yields another strong indicator that \gls{lrp} is, among the compared criteria, most capable at preserving the relevant core of the learned network function, and to dismiss unimportant parts of the model during pruning.
        }

    \edited{}{
   The strong results of \gls{lrp}, and the partial similarity between the results on the training datasets between \gls{lrp} and weight raises the question where and how both metrics (and Taylor and gradient) deviate, as it can be expected that both metrics at least select highly overlapping sets of \editedtwo{neurons and filters}{network units} for pruning and preservation.
   %, and how both metrics related to the gradient- and taylor-based pruning approaches.
    We therefore investigate in all three toy settings --- across the different number of reference samples and random seeds --- the (dis)similarities and (in)consistencies in neuron selection and ranking
    %within and between criteria
    by measuring
    %the ranking of neurons using the Spearman correlation coefficient (over a ranking of all 3000 hidden neurons), as well as
    the
    set \editedtwo{similaries}{similarities} $(S_1 \cap S_2)/\min(|S_1|,|S_2|)$ of the $k$ neurons selected for pruning (ranked \emph{first}) and preservation (ranked \emph{last}) between and within criteria.
    Since the weight criterion is not influenced by the choice of reference samples for computation, it is expected that the resulting neuron order is perfectly consistent with itself in all settings (cf.\ Table~\ref{tab:selfrank_toy}).
    What is unexpected however,
    given the results in \editedtwo{}{Figure~\ref{plot:toy_data_experiment_influence_of_n} and Figure~\ref{fig:compareresult_toy}} indicating similar model behavior after pruning to be expected between \gls{lrp}- and weight-based criteria, at least on the training data,
    is the
    \emph{minimal}
    %\emph{negative} neuron rank correlation resulting between both measures, as shown in Table~\ref{tab:lrp-vs-others-rank_toy}.
    %Similarly,
    set overlap between \gls{lrp} and weight, given the higher set similarities between \gls{lrp} and the gradient and Taylor criteria, as shown in Table~\ref{tab:lrp-vs-others-rank_toy}.
    Overall, the set overlap between the neurons ranked in the extremes of the orderings
    show that \gls{lrp}-derived pruning strategies have very little in common with the ones originating from the other criteria.
    This observation can also be made on more complex networks at hand of Figure~\ref{plot:result_remain_filter}, as shown and discussed later in this Section.
    }
       \begin{table}[!ht]
        \begin{center}
            % CASE 3 results on github
            \caption{\edited{}{Similarity analysis of neuron selection between \gls{lrp} and the other criteria, computed over 50 different random seeds.
            Higher values indicate higher similarity in neuron selection of the first/last $k$ neurons for pruning compared to \gls{lrp}.
            Note that below table reports results only for $n=10$ reference samples for criteria computation \editedtwo{}{(\underline{W}eight,  \underline{T}aylor, \underline{G}radient and \underline{L}RP)} and $k=250$ and $k=1000$.
            Similar observations have been made for $n \in [1,\editedtwo{}{2,}5,20,50,100,200]$ and $k \in [125,500]$ and can be found on \href{https://github.com/seulkiyeom/LRP_Pruning_toy_example}{github}$^3$.
            }}
            \label{tab:lrp-vs-others-rank_toy}
    		\setlength{\tabcolsep}{4pt}
    		\renewcommand{\arraystretch}{.55}
    		\scriptsize %scriptsize instead of resizebox
    		%\resizebox{\linewidth}{!}{
        		\begin{tabular}{l|cccc|cccc|cccc|cccc}
                \toprule
                \textbf{Dataset}   &   \multicolumn{4}{c|}{\textbf{first-250}}                  & \multicolumn{4}{c|}{\textbf{last-250}}                 &   \multicolumn{4}{c|}{\textbf{first-1000}}        & \multicolumn{4}{c}{\textbf{last-1000}}\\
                                        \cmidrule{2-5}                                         \cmidrule{6-9}                                           \cmidrule{10-13} \cmidrule{14-17}
                                    &   W       &   T       &   G       &   L                   &   W       &   T       &   G       &   L               &   W       &   T       &   G       &   L           &   W       &   T       &   G       &   L\\
                moon                &   0.002   &   0.006   &   0.006   &   1.000               &   0.083   &   0.361   &   0.369   &   1.000           &   0.381   &   0.639   &   0.626   &   1.000       &   0.409   &   0.648   &   0.530   &   1.000\\
                circle              &   0.033   &   0.096   &   0.096   &   1.000               &   0.086   &   0.389   &   0.405   &   1.000           &   0.424   &   0.670   &   0.627   &   1.000       &   0.409   &   0.623   &   0.580   &   1.000\\
                mult                &   0.098   &   0.220   &   0.215   &   1.000               &   0.232   &   0.312   &   0.299   &   1.000           &   0.246   &   0.217   &   0.243   &   1.000       &   0.367   &   0.528   &   0.545   &   1.000\\
                \bottomrule
                \end{tabular}
            %} %closing bracket for \resizebox
        \end{center}
    \end{table}

    	%\begin{figure}[ht!]
		%\begin{center}
	    %		\centerline{\includegraphics[width=0.85\linewidth]{figures/combined-processed-0.3.pdf}}
		%	\caption{\edited{}{\textbf{!!Alternative to Fig~\ref{plot:toy_data_experiment_influence_of_n}!!} %Pruning performance comparison of criteria depending on the number of reference samples per class used for criterion computation. 1st row: Model evaluation on the training data. 2nd row: Model evaluation on an unseen test dataset with added gaussian noise ($\mathcal{N}(0,0.3)$), which have not been used for the computation of pruning criteria. Columns: Results over different datasets. Solid lines show the average post-pruning performance of the models pruned w.r.t. to the evaluated criteria "weight" (black), "taylor" (blue), "grad(ient)" (green) and "lrp" (red) over 50 repetitions of the experiment. The dashed black line indicates the model's evaluation performance without pruning. Shaded areas around the lines show the standard deviation over the repetition of experiments.}
		%	}
		%	\label{plot:toy_data_experiment_influence_of_n_noisy}
		%\end{center}
	%\end{figure}

    %different line of analysis
    \edited{}{
    Table~\ref{tab:selfrank_toy} reports the self-similarity in neuron selection in the extremes of the ranking across random seeds (and thus sets of reference samples), for all criteria and toy settings.
    While \gls{lrp} yields a high consistency in neuron selection for \emph{both} the pruning (\emph{first-$k$}) and the preservation (\emph{last-$k$}) of neural network \editedtwo{element}{unit}s, both gradient and moreso Taylor exhibit lower self-similarities.
    The lower consistency of both latter criteria in the model components ranked last (i.e.\ preserved in the model the longest during pruning) yields an explanation for the large variation in results observed earlier: although gradient and Taylor are highly consistent in the \emph{removal} of neurons rated as  irrelevant, their volatility in the preservation of neurons which constitute the \emph{functional core} of the network after pruning yields dissimilarities in the resulting predictor function.
    The high consistency reported for \gls{lrp} in terms of neuron sets selected for pruning \emph{and} preservation, given the relatively low Spearman correlation coefficient points out only minor local perturbations of the pruning order due to the selection of reference samples.
    We find a direct correspondence between the here reported (in)consistency of pruning behavior for the three data-dependent criteria, and the in~\cite{montavon2018methods} observed ``explanation continuity'' observed for \gls{lrp} (and \emph{dis}continuity for gradient and Taylor) in neural networks containing the commonly used ReLU activation function,
    which provides an explanation for the high pruning consistency obtained with \gls{lrp}, and the extreme volatility for gradient and Taylor.
    }
    \begin{table}[!ht]
        \begin{center}
            % CASE 1 results on github
            \caption{\edited{}{A consistency comparison of neuron selection and ranking for network pruning \editedtwo{}{with criteria (\underline{W}eight,  \underline{T}aylor, \underline{G}radient and \underline{L}RP)}, averaged over all 1225 unique random seed combinations.
            Higher values indicate higher consistency in selecting the same sets of neurons and generating neuron rankings for different sets of reference samples. We report results for $n=10$ reference samples and $k=250$.
            Observations for $n \in [1,\editedtwo{}{2,}5,20,50,100,200]$ and $k \in [125,500,1000]$ are available on \href{https://github.com/seulkiyeom/LRP_Pruning_toy_example}{github}$^3$.
            }}
            \label{tab:selfrank_toy}
    		\setlength{\tabcolsep}{4pt}
    		\renewcommand{\arraystretch}{.55}
    		\scriptsize %scriptsize instead of resizebox
    		%\resizebox{\linewidth}{!}{
        		\begin{tabular}{l|cccc|cccc|cccc}
                \toprule
                \textbf{Dataset}    & \multicolumn{4}{c|}{\textbf{first-250}}                       & \multicolumn{4}{c|}{\textbf{last-250}}                &   \multicolumn{4}{c}{\textbf{Spearman Correlation}}\\
                                        \cmidrule{2-5}                                              \cmidrule{6-9}                                          \cmidrule{10-13}
                                    &   W       &   T       &   G       &   L                       &   W           &   T       &   G       &   L           &   W       &   T       &   G       &   L\\
                moon                &   1.000   &   0.920   &   0.918   &   0.946                   &   1.000       &   0.508   &   0.685   &   0.926       &   1.000   &   0.072   &   0.146   &   0.152\\
                circle              &   1.000   &   0.861   &   0.861   &   0.840                   &   1.000       &   0.483   &   0.635   &   0.936       &   1.000   &   0.074   &   0.098   &   0.137\\
                mult                &   1.000   &   0.827   &   0.829   &   0.786                   &   1.000       &   0.463   &   0.755   &   0.941       &   1.000   &   0.080   &   0.131   &   0.155\\
                \bottomrule
                \end{tabular}
            %} %closing bracket for \resizebox
        \end{center}
    \end{table}
    \edited{}
    {A supplementary analysis of the neuron selection consistency of \gls{lrp} over different counts of reference samples $n$, demonstrating the requirement of only very few reference samples per class in order to obtain stable pruning results, can be found in Supplementary Results 1.
    }

    \edited{}{
    Taken together, the results of Tables~\ref{tab:lrp-vs-others-rank_toy} to~\ref{tab:selfrank_toy} and \editedtwo{}{Supplementary Tables 1 and 2} elucidate that \gls{lrp} constitutes --- compared to the other methods --- an \emph{orthogonal} pruning criterion which is very consistent in its selection of (un)important neural network \editedtwo{element}{unit}s, while remaining adaptive to the selection of reference samples for criterion computation.
    Especially the similarity in post-pruning model performance to the \emph{static} weight criterion indicates that both metrics are able to find valid, yet completely different pruning solutions.
    However, since \gls{lrp} can still benefit from the influence of reference samples,
    we will show in Section~\ref{sec:no_finetuning} that our proposed criterion is able to outperform not only weight, but all other criteria in Scenario 2, where pruning is is used instead of fine-tuning as a means of domain adaptation. This will be discussed in the following sections.

    %From this we conclude that the use of \gls{lrp} as a pruning criterion might find feasible application in context of domain adaptation tasks,  with pruning as a fine-tuning step, which will be investigated in following sections.
    }

    %despite high consistency of lrp-results across N and seeds, the neurons selected for pruning have little in common with the other criteria, esp. weight. table here. means: both weight and lrp find good solutions, yet completely different ones.

    %picking up minor perturbations for lrp: number of L does pretty much not affect
    \edited{
    These results show that among the considered pruning criteria, pruning with the relevance-criterion based on \gls{lrp} best preserves the \emph{effective} core embodied by the example models,
    and thus prediction performance in every toy setting.
    In contrast to the other heuristic\edited{s-based}{} pruning criteria, \gls{lrp} directly relates relevance to the classification output, hence allows to safely remove the unimportant (w.r.t.~classification) \editedtwo{element}{unit}s.
    Furthermore, Figure~\ref{plot:toy_data_experiment} shows that when pruning 1000 out of 3000 neurons,
    \gls{lrp}-based pruning results in only minimal change in the decision boundary, compared to the other criteria.
    %There are some distortions around borders between two classes.
    One can also clearly see that compared to weight- and \gls{lrp}-based criteria, models pruned by gradient-based criteria misclassify a large part of samples.
    }{}

	\subsection{Pruning Deep Image Classifiers for Large-scale Benchmark Data}
	%Data and Model description
	\edited{The performance}{We now evaluate the performance of all} \edited{of the}{}pruning criteria \edited{is evaluated}{}on the \glspl{cnn}, VGG-16\edited{and}{,} AlexNet \edited{}{as well as ResNet-18 and ResNet-50, --- popular models in compression research~\cite{WangZWH18} --- all of which are} pre-trained on ILSVRC~2012 \edited{}{(ImageNet)}\edited{that are popular in model compression research~\cite{WangZWH18}}{}.
	VGG-16 consists of 13 convolutional layers with 4224 filters and 3 fully-connected layers and AlexNet contains 5 convolutional layers with 1552 filters and 3 fully-connected layers. In dense layers, there exist \edited{8192 + N(\# of classes) of}{4,096+4,096+$k$} neurons \edited{}{(i.e.\ filters)}, respectively\edited{}{, where $k$ is the number of output classes}.
	In terms of complexity of the model, the pre-trained VGG-16 and AlexNet on ImageNet originally consist of 138.36/60.97 million of parameters and 154.7/7.27 \edited{}{\gls{gmacs}} \edited{for}{(as a measure of \gls{flop})}, respectively.
	%\edited{}{(values obtained via tools by~\citet{pakhomov2019deep})}.
	%~\cite{HasanPourRVS16}.
	\edited{}{ResNet-18 and ResNet-50 consist of 20/53 convolutional layers with 4,800/26,560 filters. In terms of complexity of the model, the pre-trained ResNet-18 and ResNet-50 on ImageNet originally consist of 11.18/23.51 million of parameters and 1.82/4.12 \gls{gmacs} (as a measure of \gls{flop}), respectively.}

	%\edited{}{\textbf{The ResNet-18... replicate above info for ResNets here.}}

	%\edited{}{Note that the experiments on ResNets use an implementation\footnote{\edited{}{\url{https://github.com/kuangliu/pytorch-cifar}}} which produces better performance on CIFAR-10.}

	\edited{}{
	Furthermore, since the \gls{lrp} scores are not implementation-invariant and depend on the \gls{lrp} rules used for the batch normalization (BN) layers, we convert a trained ResNet into a canonized version, which yields the same predictions up to numerical errors. The canonization fuses a sequence of a convolution and a BN layer into a convolution layer with updated weights\footnote{\edited{}{See \texttt{bnafterconv\_overwrite\_intoconv(conv,bn)} in the file \texttt{lrp\_general6.py} in \url{https://github.com/AlexBinder/LRP_Pytorch_Resnets_Densenet} }} and resets the BN layer to be the identity function.
	This removes the BN layer effectively by rewriting a sequence of two affine mappings into one updated affine mapping~\edited{}{\cite{guillemot2020breaking}}.
	The second change replaced calls to {\tt torch.nn.functional} methods and the summation  in the residual connection by classes derived from \texttt{torch.nn.Module} which then were wrapped by calls to \texttt{torch.autograd.function} to enable custom backward computations suitable for \gls{lrp} rule computations.}
	%\edited{}{\begin{align*}\text{conv-layer: }y &= w_{conv}\cdot x + b_{conv,c}  \\ \text{bn-layer: } z&= w_c \,(y - \mu_{bn})/s_c + bn_c\\ \text{updated conv-layer: } z&= (w_c/s_c) w_{conv} \cdot x + \,(b_{conv,c} - \mu_{bn})/s_c + bn_c\end{align*}}

	%\edited{}{\textbf{Note however, ...}  \textbf{TODO ADD: model canonization of ResNet, whcih is not required for VGG. Cite hui and binder 2018: LRP is not implementation invariant. minor drawback, model can be brought in to a functionally equivalent, "right form". BN is static during inference. problem for fine-tuning? ad-hoc creation of "ready for pruning" resnet always possible.}}

	Experiments are performed within the \textit{PyTorch} and \textit{torchvision} frameworks under \textit{Intel(R) Xeon(R) CPU E5-2660 2.20GHz} and \textit{NVIDIA Tesla P100 with 12GB} for GPU processing.
	We evaluated the criteria on six public datasets (Scene 15~\cite{lazebnik2006beyond}, Event 8, Cats and Dogs~\cite{asirra2007}, Oxford Flower 102~\cite{nilsback2008automated}, C\edited{ifar 10}{IFAR-10}, and ILSVRC~2012~\cite{ILSVRC15}). For more detail on the datasets and the preprocessing, see Supplementary Methods 1.
	\editedtwo{}{\textit{Our complete experimental setup covering these datasets is publicly available at
	\href{https://github.com/seulkiyeom/LRP_pruning}{\texttt{https://github.com/seulkiyeom/LRP$\_$pruning}}.}}

	\begin{figure}[h!]
		\vskip 0.2in
		\begin{center}
			\centerline{\includegraphics[width=0.99\linewidth]{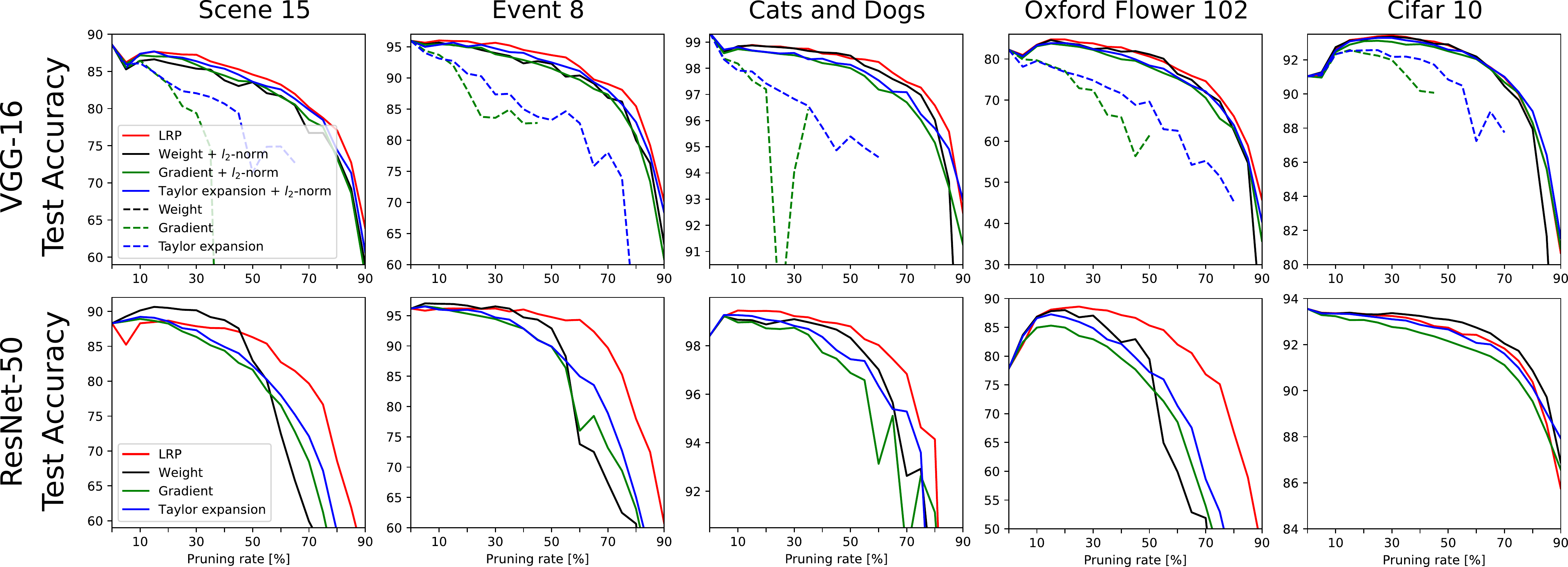}}
			\caption{Comparison of test accuracy in different criteria as pruning rate increases on VGG-16 (top) and ResNet-50 (bottom) with five datasets.
			\edited{}{Pruning \emph{with} fine-tuning. Prematurely terminated lines in above row of panels indicate that during pruning, the respective criterion removed filters vital to the network structure  by disconnecting the model input from the output.}
			}
			\label{plot:result_vgg16}
		\end{center}
		\vskip -0.2in
	\end{figure}

	\begin{table}[!ht]
        \begin{center}
            \caption{
            \edited{}{
            A performance comparison between criteria (\underline{W}eight, \underline{T}aylor, \underline{G}radient with $\ell_2$-norm  each and \underline{L}RP) and the \underline{U}npruned model for \textbf{VGG-16} (top) and \textbf{ResNet-50} (bottom)
            on five different image benchmark datasets.
            Criteria are evaluated at fixed pruning rates per model and dataset, identified as \texttt{$\langle$dataset$\rangle$@$\langle$percent\_pruned\_filters$\rangle$\%}.
            We report test accuracy (in \%), (training) loss ($\times 10^{-2}$), number of remaining parameters ($\times 10^7)$ and FLOPs (in GMAC) per forward pass.
            For all measures except accuracy, lower outcomes are better.
            }}
            \label{tab:compareresult_vgg16_resnet50}
    		\setlength{\tabcolsep}{4pt}
    		\renewcommand{\arraystretch}{.55}
    		\scriptsize %scriptsize AND resizebox for this one
    		\resizebox{\linewidth}{!}{
        		\begin{tabular}{l|rrrrr|rrrrr|rrrrr}
                \toprule
                \textbf{VGG-16}         &   \multicolumn{5}{c|}{\textbf{Scene 15 @ 55\%}}                                           & \multicolumn{5}{c|}{\textbf{Event 8 @ 55\%}}                                                      & \multicolumn{5}{c}{\textbf{Cats \& Dogs @ 60\%}}                                  \\
                                            \cmidrule{2-6}                                                                            \cmidrule{7-11}                                                                                     \cmidrule{12-16}
                                        &   U           &   W       &  T                &   G           &   L                       &   U           &   W       &  T                &   G               &   L                           &   U           &   W       &  T          &   G             &   L                   \\
                Loss                    &   2.09        &   2.27    & 1.76              &   1.90        &  \textbf{1.62}            &   0.85        &   1.35    &  1.01             &   1.18            &   \textbf{0.83}               &   0.19        &   0.50    &  0.51       &   0.57          &   \textbf{0.44}       \\
                Accuracy                &   88.59       &   82.07   & 83.00             &   82.72       &  \textbf{83.99}           &   95.95       &   90.19   &  91.79            &   90.55           &   \textbf{93.29}              &   99.36       &   97.90   &  97.54      &   97.19         &   \textbf{98.24}      \\
                Params                  &   119.61      &   56.17   & 53.10             &   53.01       &  \textbf{49.67}           &   119.58      &   56.78   &  48.48            &   50.25           &   \textbf{47.35}              &   119.55      &   47.47   &  51.19      &   57.27         &   \textbf{43.75}      \\
                FLOPs                   &   15.50       &   8.03    & \textbf{4.66}     &   4.81        &  6.94                     &   15.50       &   8.10    &  5.21             &   \textbf{5.05}   &   7.57                        &   15.50       &   7.02    &  3.86       &   \textbf{3.68} &   6.49                \\
                \multicolumn{1}{c}{}    & \multicolumn{5}{c}{}                                                                      & \multicolumn{5}{c}{}                                                                              & \multicolumn{5}{c}{}                                                          \\
                                        & \multicolumn{5}{c|}{\textbf{Oxford Flower 102 @ 70\%}}                        &   \multicolumn{5}{c|}{\textbf{CIFAR-10 @ 30\%}}     & \multicolumn{5}{c}{}\\
                                        \cmidrule{2-6}                                                                      \cmidrule{7-11}
                                        &   U       &   W       &  T            &   G       &   L                       &   U       &   W           &  T            &   G       &   L               &      &      &      &      &      \\
                Loss                    &   3.69    &   3.83    &  3.27         &   3.54    &  \textbf{2.96}            &   1.57    &   1.83        &  1.76         &   1.80    & \textbf{1.71}     &      &      &      &      &      \\
                Accuracy                &   82.26   &   71.84   &  72.11        &   70.53   &  \textbf{74.59}           &   91.04   &   93.36       &  93.29        &   93.05   & \textbf{93.42}    &      &      &      &      &      \\
                Params                  &   119.96  &   39.34   &  41.37        &   42.68   &  \textbf{37.54}           &   119.59  & \textbf{74.55}&  97.30        &   97.33   &   89.20           &      &      &      &      &      \\
                FLOPs                   &   15.50   &   5.48    & \textbf{2.38} &   2.45    &  4.50                     &   15.50   &   11.70       &  \textbf{8.14}&   8.24    &   9.93            &      &      &      &      &      \\
                \bottomrule
                % ABOVE: VGG results. Below: ResNet-50 results
                \toprule
                \textbf{ResNet-50}         &   \multicolumn{5}{c|}{\textbf{Scene 15 @ 55\%}}                                           & \multicolumn{5}{c|}{\textbf{Event 8 @ 55\%}}                                                      & \multicolumn{5}{c}{\textbf{Cats \& Dogs @ 60\%}}                                  \\
                                            \cmidrule{2-6}                                                                            \cmidrule{7-11}                                                                                     \cmidrule{12-16}
                                        &   U           &   W       &  T                &   G           &   L                       &   U           &   W       &  T                &   G               &   L                           &   U           &   W       &  T          &   G             &   L                   \\
                Loss                    &   0.81           &   1.32       &  1.08                &   1.32           &   \textbf{0.50}                       &   0.33           &   1.07       &  0.63                &   0.85               &   \textbf{0.28}                           &   0.01           &   0.05       &  0.06          &   0.21             &   \textbf{0.02}                   \\
                Accuracy                &   88.28           &   80.17      &  80.26                &   78.71           &   \textbf{85.38}                       &   96.17           &   88.27       &  87.55                &   86.38               &   \textbf{94.22}                           &   98.42           &   97.02       &  96.33          &   93.13             &   \textbf{98.03}                   \\
                Params                  &   23.54           &   14.65       &  12.12                &   \textbf{11.84}           &   13.73                       &   23.52           &   13.53       &  \textbf{11.85}                &   11.93               &   14.05                           &   23.51           &   12.11       &  \textbf{10.40}          &   10.52             &   12.48                   \\
                FLOPs                   &   4.12           &   3.22       &  2.45                &   \textbf{2.42}           &   3.01                       &   4.12           &   3.16       &  2.48               &   \textbf{2.47}               &   3.10                           &   4.12           &   3.04       &  2.40          &   \textbf{2.27}             &   2.89                   \\
                \multicolumn{1}{c}{}    & \multicolumn{5}{c}{}                                                                      & \multicolumn{5}{c}{}                                                                              & \multicolumn{5}{c}{}                                                          \\
                                        & \multicolumn{5}{c|}{\textbf{Oxford Flower 102 @ 70\%}}                        &   \multicolumn{5}{c|}{\textbf{CIFAR-10 @ 30\%}}     & \multicolumn{5}{c}{}\\
                                        \cmidrule{2-6}                                                                      \cmidrule{7-11}
                                        &   U       &   W       &  T            &   G       &   L                       &   U       &   W           &  T            &   G       &   L               &      &      &      &      &      \\
                Loss                    &   0.82       &   3.04       &  2.18            &   2.69       &   \textbf{0.83}                       &   0.003       &   \textbf{0.002}           &  0.004            &   0.009       &   0.003               &      &      &      &      &      \\
                Accuracy                &   77.82       &   51.88       &  58.62            &   53.96       &   \textbf{76.83}                       &   93.55       &   \textbf{93.37}           &  93.15            &   92.76       &   93.23               &      &      &      &      &      \\
                Params                  &   23.72       &   9.24       &  8.82            &   \textbf{8.48}       &   9.32                       &   23.52       &   19.29           &  \textbf{18.10}            &   17.96       &   18.11               &      &      &      &      &      \\
                FLOPs                   &   4.12       &   2.55       &  \textbf{1.78}            &   1.81       &   2.38                       &   1.30       &   1.14           &  1.06            &   \textbf{1.05}       &   1.16               &      &      &      &      &      \\
                \bottomrule
                \end{tabular}
            } %closing bracket for \resizebox
        \end{center}
    \end{table}

	%Implementation environment
	% Training details
	\edited{We}{In order to prepare the models for evaluation, we first} fine-tuned the model\edited{}{s} for 200 epochs with constant learning rate 0.001 and batch size of 20. We used the Stochastic Gradient Descent (SGD) optimizer with momentum of 0.9.
	In addition, we also apply dropout to the fully-connected layers with probability of 0.5. Fine-tuning and pruning are performed on the training set, while results are evaluated on each test dataset.
	Throughout the experiments, we iteratively prune 5\% of all the filters in the network by eliminating \editedtwo{neurons}{units} including their input and output connections.
	In Scenario~1, we subsequently fine-tune and re-evaluate the model to account for dependency across parameters and regain performance, as it is common.
% 	When filters are iteratively pruned, a new pruned model is re-created and the remaining parameters of the modified layers as well as the unaffected layers are copied into the new pruned model.

	%Results are here
	%Our method prunes the less useful convolutional filters from a well-trained model for computational efficiency while minimizing the accuracy drop. In the previous approaches, they measure the relative importance of a filter in each layer by calculating the sum or the average of its absolute criteria using additional regularization such as $l_2$-norm for global rescaling. However, due to the conservative property of LRP, it can automatically measure the globally relative importance of filters without further regularization.

	\begin{figure}[ht!]
		\vskip 0.2in
		\begin{center}
			\centerline{\includegraphics[width=0.99\linewidth]{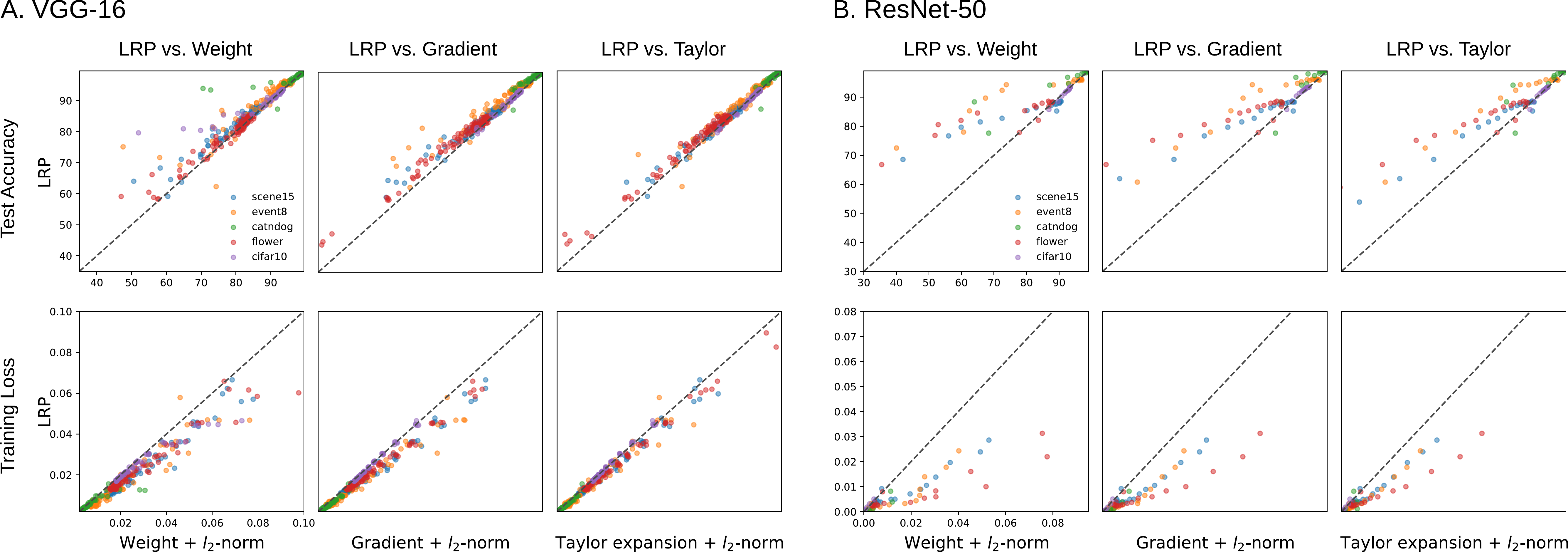}}
			\caption{Performance comparison of the proposed method (i.e.\ \gls{lrp}) and other criteria on VGG-16 \edited{}{and ResNet-50} with five datasets.
			\edited{}{Each point in the scatter plot corresponds to the performance at a specific pruning rate of two criteria, where the vertical axis shows the performance of our \gls{lrp} criterion and the horizontal axis the performance of a single other criterion (compare to Figure~\ref{plot:result_vgg16} that displays the same data for more than two criteria). The black dashed line shows the set of points where models pruned by one of the compared criteria would exhibit identical performance to \gls{lrp}. For accuracy, higher values are better. For loss, lower values are better.}}
			\label{plot:perf_compare_vgg16}
		\end{center}
		\vskip -0.2in
	\end{figure}

	\subsubsection{Scenario 1: Pruning with Fine-tuning} %Table 2,3 Figure 2
	\label{sec:experiments/with_finetuning}
	On the first scenario,
	we retrain the model after each iteration of pruning in order to regain lost performance.
	We then evaluate the performance of the different pruning criteria after each pruning-retraining-step.
	\begin{figure*}[ht!]
		\vskip 0.2in
		\begin{center}
			\centerline{\includegraphics[width=\linewidth]{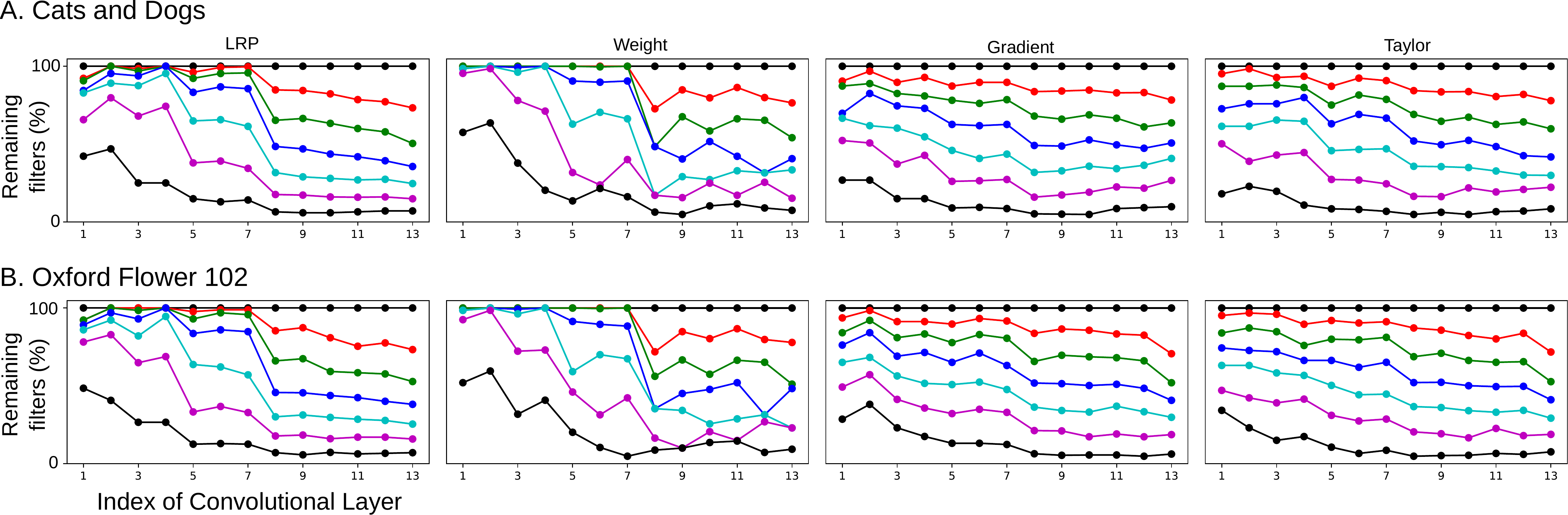}}
			\caption{\edited{
			Comparison of the ratio of remaining convolutional filters at different depths of the network for different pruning rates (0 (black), 15 (red), 30 (green), 45 (blue), 60 (cyan), 75 (magenta), and 90\% (black) from top to bottom) across criteria.
			}{
			An observation of per-layer pruning performed w.r.t\ the different evaluated criteria on VGG-16 and two datasets.
			Each colored line corresponds to a specific (global) ratio of filters pruned from the network
			({\color{black} \bf black (top) : 0\%},
			{\color{red} \bf red : 15\%},
			{\color{OliveGreen} \bf green: 30\%},
			{\color{blue} \bf blue: 45\%},
			{\color{Plum} \bf violet: 75\%} and
			{\color{black} \bf black (bottom) 90\%}).
			The dots on each line identify the ratio of pruning applied to specific convolutional layers, given a global ratio of pruning, depending on the pruning criterion.}}
			\label{plot:result_remain_filter}
		\end{center}
		\vskip -0.2in
	\end{figure*}
	That is,
	we quantify the importance of each filter by the magnitude of the respective criterion and iteratively prune \edited{the}{} 5\% of \edited{}{all} filters (w.r.t. the original number of filters in the model) rated least important in each pruning step.
	Then,
	we compute and record the training loss, test accuracy, number of remaining parameters and total estimated \gls{flop}.
	We assume that the least important filters should have only little influence on the prediction and thus incur the lowest performance drop if they are removed from the network.
	%In the following experiments, only the convolutional filters are pruned.

	Figure~\ref{plot:result_vgg16} (and Supplementary Figure~2) depict test accuracies \edited{as}{with} increasing \edited{the}{}pruning rate in VGG-16 \edited{}{and ResNet-50 (and AlexNet and ResNet-18, respectively)} after fine-tuning for each dataset and each criterion.
	\edited{At each pruning iteration, we remove 5\% of the weights/filters in the entire network based on the magnitude of the different criteria.}{}
	It is observed that \gls{lrp} achieves higher test accuracies
	\edited{ as well as the lower training losses compared to other criteria for the entire range of pruning iterations on every dataset}{compared to other criteria in a large majority of cases} (see Figure~\ref{plot:perf_compare_vgg16} and Supplementary Figure~1).
	These results demonstrate that the performance of \gls{lrp}-based pruning is stable and independent of the chosen dataset.
	% due to the inherent conservation property.
	Apart from performance, regularization by layer \edited{must be}{is} a critical constraint which obstructs the expansion \edited{}{of some} of the criteri\edited{on}{a} toward several pruning strategies such as local pruning, global pruning, etc.
	Except for the \gls{lrp} criterion, all criteria perform substantially worse without $l_p$ regularization compared to those with $l_p$ regularization and result in unexpected interruptions during the pruning process due to the biased redistribution of importance in the network~(cf. top rows of Figure~\ref{plot:result_vgg16} and Supplementary Figure~2).

    % VGG table results
	Table~\ref{tab:compareresult_vgg16_resnet50} shows the predictive performance of the different criteria in terms of training loss, test accuracy, number of remaining parameters and \gls{flop}\edited{}{, for the VGG-16 and ResNet-50 models. Similar results for AlexNet and ResNet-18 can be found in Supplementary Table~2}.
	Except for C\edited{ifar10}{IFAR-10}, the highest compression rate (i.e.\ lowest number of parameters) could be achieved by the proposed \gls{lrp}-based criterion (\edited{column "\#Parameters" in Table~\ref{tab:compareresult_vgg16_resnet50}}{row ``Params''}) \edited{}{for VGG-16, but not for ResNet-50}.
	However, in terms of \gls{flop}, the proposed criterion only outperformed the weight criterion, but not the Taylor and Gradient criteria (\edited{see column "\gls{flop}" in Table~\ref{tab:compareresult_vgg16_resnet50}}{row``FLOPs''}).
	This is due to the fact that a reduction in number of \gls{flop} depends on the location where pruning is applied within the network:
	Figure~\ref{plot:result_remain_filter} shows that the \gls{lrp} and weight criteria focus the pruning on upper layers closer to the model output, whereas the Taylor and Gradient criteria focus more on the lower layers.

	% Initial performance increase
	Throughout the pruning process usually a gradual decrease in performance can be observed.
	However, with the Event 8, Oxford Flower 102\edited{, Cifar 10}{ and CIFAR-10} datasets, pruning leads to an initial performance increase, until a pruning rate of approx.\ 30\% is reached.
	This behavior has been reported before in the literature and might stem from improvements of the model structure through elimination of filters related to classes in the source dataset (i.e.,~ILSVRC~2012) that are not present in the target dataset anymore~\cite{LiuW017}.
    %
	% Alexnet results
	\editedtwo{}{Supplementary Table~3} and Supplementary Figure~2 similarly show that \gls{lrp} achieves the highest test accuracy in AlexNet and ResNet-18  for nearly all pruning ratios with almost every dataset.
	%Furthermore, as discussed in Section~\ref{methods}, due to the widespread use of \gls{lrp}, we do not take into account dataset and model type, which could be a powerful advantage going forward.
	%To measure the computation complexity, a widely used metric is the number of float-point operations, or FLOPs . However, FLOPs is an indirect metric. It is an approximation of, but usually not equivalent to the direct metric that we really care about, such as speed or latency. 이유는  several important factors that have considerable affection on speed are not taken into account by FLOPs.

	%50\%(맞나?)까지는 초기 성능이 유지된다 (model can drop at least 50\% of convolutional filters without affecting the accuracy drop)

	%\subsubsection*{Transition of the Number of the Convolutional Filter}
% 	Pruning layers across the network gives a holistic view of the robustness of a smaller network.
    Figure~\ref{plot:result_remain_filter} shows the number of the remaining convolutional filters for each iteration. We observe that, on the one hand, as pruning rate increases, the convolutional filters in earlier layers that are associated with very generic features, such as edge and blob detectors, tend to generally be preserved as opposed to those in latter layers which are associated with abstract, task-specific features.
    % TODO is it wise to call LRP a weight-based criterion? Goes against our punchline that weights (weights ALONE) are not meaningful
    On the other hand, the \gls{lrp}- and weight-criterion first keep the filters in early layers in the beginning, but later aggressively prune filters near the input which now have lost functionality as input to later layers, compared to the gradient-based criteria such as gradient and Taylor-based approaches.
    Although gradient-based criteria also adopt the greedy layer-by-layer approach, we can see that gradient-based criteria pruned the less important filters almost uniformly across all the layers due to re-normalization of the criterion in each iteration.
    However, this result contrasts with previous gradient-based works~\cite{MolchanovTKAK16, sun2017meprop} that have shown that \editedtwo{neurons}{units} deemed unimportant in earlier layers, contribute significantly compared to \editedtwo{neurons}{units} deemed important in latter layers.
    In contrast to this, \gls{lrp} can efficiently preserve \editedtwo{neurons}{units} in the early layers --- as long as they serve a purpose --- despite of iterative global pruning.

	\subsubsection{Scenario 2: Pruning without Fine-tuning}
	\label{sec:no_finetuning}
% 	Pruning and retraining of the network in a layer-by-layer fashion can be very time-consuming. In this respect, we still have a question:
	In this section, we evaluate whether pruning works well if only a (very) limited number of samples is available for quantifying the pruning criteria.
	To the best of our knowledge, there are no previous studies that show the performance of pruning approaches when acting w.r.t. very small amounts of data.
	With large amounts of data available (and even though we can expect reasonable performance after pruning),
	an iterative pruning and fine-tuning procedure of the network
	can amount to a very time consuming and computationally heavy process.
	From a practical point of view, this issue becomes a significant problem,
	e.g. with limited computational resources (mobile devices or in general; consumer-level hardware) and reference data (\edited{}{e.g.,} private photo collections),
	where capable and effective one-shot pruning approaches are desired
	and only little leeway (or none at all) for \edited{post-pruning fine-tuning strategies}{fine-tuning strategies after pruning} is available.

	To investigate whether pruning is possible also in these scenarios, we performed experiments with a relatively small number of data on the 1) Cats \& Dogs and 2) \edited{}{subsets from the} \edited{ILVSRC}{ILSVRC}~2012 \edited{datasets}{classes}.
	% Cats and Dogs
    On the Cats \& Dogs dataset, we only used 10 samples each from the ``cat'' and ``dog'' classes to prune the (on ImageNet) pre-trained \edited{}{AlexNet,} VGG-16\edited{}{, ResNet-18 and ResNet-50} network\edited{}{s} with the goal of domain/dataset adaption.
    The binary classification (i.e. ``cat'' vs. ``dog'') is a subtask \edited{of}{within the} ImageNet \edited{}{taxonomy} and corresponding output neurons can be identified by its WordNet\footnote{\url{http://www.image-net.org/archive/wordnet.is\_a.txt}} associations.
    \edited{}{This experiment implements the task of domain adaptation.}

    % ImageNet
    \edited{O}{In a second experiment o}n the ILSVRC~2012 dataset, we randomly chose \mbox{$k=3$} classes for \edited{}{the task of} model specialization, selected only $n=10$ images per class from the training set and used them to compare the different pruning criteria.
    For each criterion, we used the same selection of classes and samples.
    In \edited{these}{both experimental settings}\edited{experiments}{},  \emph{we do not fine-tune} the models after each pruning iteration, in contrast to \edited{}{\emph{Scenario 1} in} Section~\ref{sec:experiments/with_finetuning}.
    \edited{}{The obtained post-pruning model } performance is averaged over 20 random selections of classes \edited{}{(ImageNet)} and samples \edited{}{(Cats \& Dogs)} to account for randomness.
    Please note that before pruning, we first\edited{ly}{} \edited{reconstructed}{restructured} \edited{}{the models' fully connected} output layers \edited{which has the size of N $\times$ k}{to only preserve the task-relevant $k$ network outputs} by eliminating the \edited{}{$1000 - k$} redundant \edited{network outputs of 1000 - $k$}{output neurons}.

	Furthermore,
	as our target datasets are relatively small and only have an extremely reduced set of target classes,
	the pruned
	%\edited{}{(sequential VGG-like)}
	models \edited{would}{could} still be very heavy w.r.t. memory requirements if the pruning process would be limited to the convolutional layers, as in Section~\ref{sec:experiments/with_finetuning}.
    More specifically, while convolutional layers dominantly constitute the source of computation cost \edited{}{(\gls{flop})}, fully connected layers are proven to be more redundant~\cite{li2016pruning}.
    In this respect, we applied pruning procedures in both fully connected layers and convolutional layers \edited{}{in combination for VGG-16}.

    For pruning, we iterate a sequence of first pruning filters from the convolutional layers, followed by a step of pruning \editedtwo{filters/}{neurons} from the model's fully connected layers.
    \edited{}{Note that both evaluated ResNet architectures mainly consist of convolutional- and pooling layers, and conclude in a single dense layer, of which the set of input neurons are only affected via their inputs by pruning the below convolutional stack.
    We therefore restrict the iterative pruning filters from the sequence of dense layers of the feed-forward architecture of the VGG-16.}

    \begin{figure}
        \centering
        \includegraphics[width=\linewidth]{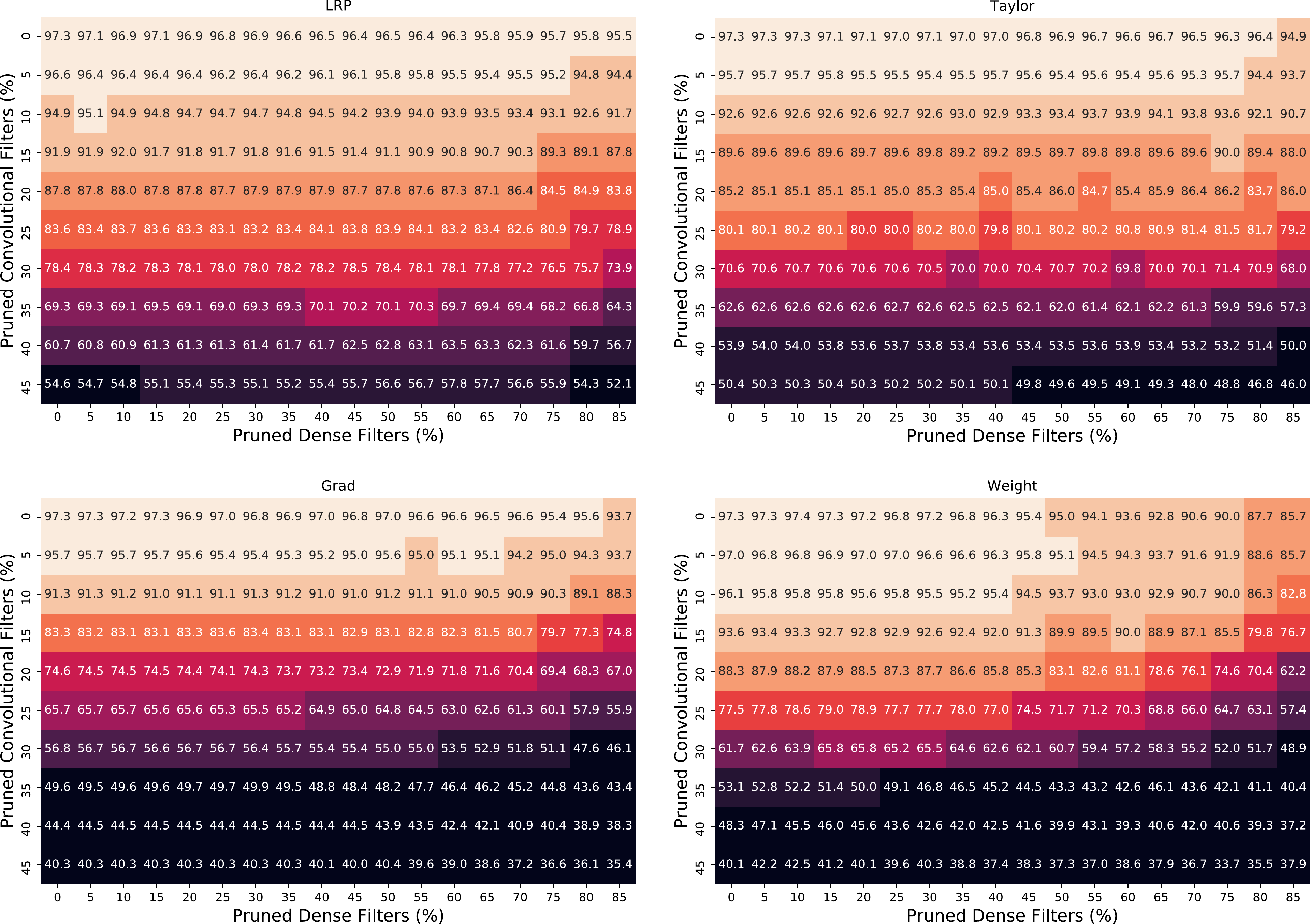}
        \caption{Test accuracy after pruning of \edited{}{$n\%$ of} convolutional \edited{}{(rows)} and \edited{}{$m\%$ of} fully connected \edited{}{(columns)} \edited{layers}{filters} on VGG-16 \emph{without} fine-tuning for a random subset of the classes from ILSVRC~2012 ($k$ = 3) based on different criteria (averaged over 20 repetitions).
        %\textit{
        \edited{}{Each color represents a range of 5\% in test accuracy. The brighter the color the better the performance after a given degree of pruning}
        %}
        .}
        \label{fig:compareresult_small}
    \end{figure}

    \begin{figure}
        \centering
        \includegraphics[width=\linewidth]{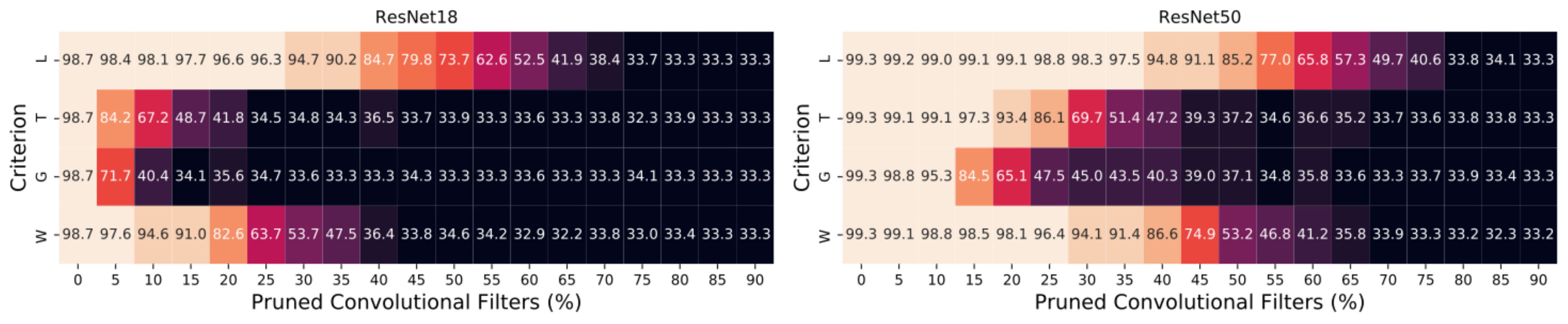}
        \caption{\edited{}{Test accuracy after pruning of \edited{}{$n\%$ of} convolutional filters on ResNet18 and ResNet50 \emph{without} fine-tuning for a random subset of the classes from ILSVRC~2012 ($k$ = 3) based on the criteria \underline{W}eight, \underline{T}aylor, \underline{G}radient with $\ell_2$-norm and \underline{L}RP (averaged over 20 repetitions).
        \textit{
        Compare to Figure~\ref{fig:compareresult_small}
        }
        .}}
        \label{fig:compareresult_small_resnet}
    \end{figure}

	%\edited{Table}{Figure}~\ref{fig:compareresult_small} indicates the \edited{performances of}{model performance after the application of} each criterion for classifying a small number of classes \mbox{($k$ = 3)} from the ILSVRC~2012 dataset \edited{}{with VGG-16}.
	The \edited{performances of}{model performance after the application of} each criterion for classifying a small number of classes \mbox{($k$ = 3)} from the ILSVRC~2012 dataset is indicated in Figure~\ref{fig:compareresult_small} for VGG~16\edited{}{ and Figure~\ref{fig:compareresult_small_resnet} for ResNets (please note again that ResNets do not have fully-connected layers)}.
    During pruning at fully-connected layers, no significant difference across different pruning ratios can be observed.
    Without further fine-tuning, pruning weights/filters at the fully connected layers can retain performance efficiently.
    However, there is a certain difference between \gls{lrp} and other criteria with increasing pruning ratio of convolutional layers
    \edited{}
    {for VGG-16/ResNet-18/ResNet-50, respectively:
    (\gls{lrp} vs. Taylor with $l_2$-norm; up to of 9.6/61.8/51.8\%,
    \gls{lrp} vs. gradient with $l_2$-norm; up to 28.0/63.6/54.5~\%,
    \gls{lrp} vs. weight with $l_2$-norm; up to 27.1/48.3/30.2~\%)}.
    Moreover, pruning convolutional layers needs to be carefully managed compared to pruning fully connected layers.
    % TODO this sentence is redundant and should be moved to the discussion
    We can observe that \gls{lrp} is applicable for pruning any layer type (i.e.\ fully connected, convolutional, pooling, etc.) efficiently.
    Additionally, as mentioned in \edited{}{Section}~\ref{sec:proposed_method}, our method can be applied to general network architectures because \edited{it}{}it can automatically measure the importance of weights or filters in a global (network-wise) context without further normalization.

\begin{figure}[ht!]
		\vskip 0.2in
		\begin{center}
			\centerline{\includegraphics[width=\linewidth]{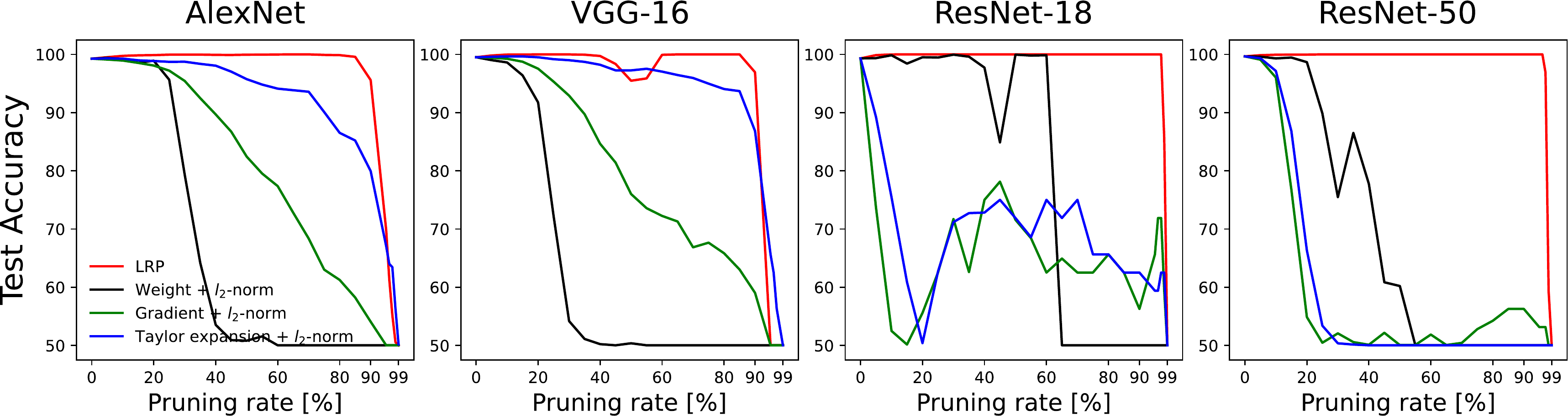}}
	        \caption{Performance comparison of pruning without fine-tuning for \edited{VGG-16}{AlexNet, VGG-16, ResNet-18 and ResNet-50} \edited{}{based on only few (10) samples per class}\edited{ with small samples}{} from \edited{}{the} Cats \& Dogs dataset, as a means for domain adaption. \editedtwo{}{Additional results on further target domains can be found in the Supplement with Supplementary Figure 3.}}
			\label{plot:result_small_sample_1}
		\end{center}
		\vskip -0.2in
	\end{figure}

    Figure~\ref{plot:result_small_sample_1} shows the test accuracy \edited{and training loss} as \edited{}{a} function of the pruning ratio,
    in context a domain adaption task \edited{}{from ImageNet} towards the Cats \& Dogs dataset \edited{}{for all models}.
    As the pruning ratio increases, we can see that even without fine-tuning, using \gls{lrp} as pruning criterion can keep the test accuracy \edited{and training loss}{} not only stable, but close to 100\%\edited{and 0 respectively}{}, given the extreme scarcity of data in this experiment.
    In contrast, the performance decreases significantly when using \edited{}{the other criteria requiring an application of} the $l_2$-norm\edited{based criteria (i.e. weight, gradient and Taylor)}{}.
    Initially, the performance is even slightly increasing when pruning with \gls{lrp}.
    During iterative pruning, unexpected changes in accuracy with \gls{lrp} (\edited{}{for 2 out of 20 repetitions of the experiment}) have been shown around 50~-~55\% pruning ratio, but accuracy is regained quickly again.
    \edited{}{However, only the VGG-16 model seems to be affected, and none other for this task. For both ResNet models, this phenomenon occurs for the other criteria instead. A series of in-depth investigations of this momentary decrease in performance did not lead to any insights and will be subject of future work\footnote{\edited{}{We consequently have to assume that this phenomenon marks the downloaded pre-trained VGG-16 model as an outlier in this respect.
    A future line of research will dedicate inquiries about the circumstances leading to intermediate loss and later recovery of model performance during pruning.}}.}

    By pruning \edited{}{over} 99\% of convolutional filters in the \edited{network}{networks} using our proposed method, we can have 1) greatly reduced computational cost, 2) faster forward and backward processing \edited{}{(e.g.\ for the purpose of further training, inference or the computation of attribution maps)}, and 3) a lighter model even in the small sample case,
    all while adapting \edited{an}{} off-the-shelf pre-trained ImageNet \edited{model}{models} towards a dog-vs.-cat classification task.

\edited{}{
\section{Discussion}
\label{Discussion}
 }

\edited{}
{
% Better pruning performance
Our experiments demonstrate that the novel \gls{lrp} criterion consistently performed well compared to other criteria across various datasets, model architectures and experimental settings, and oftentimes outperformed the competing criteria.
This is especially pronounced in our Scenario 2 (cf.~Section~\ref{sec:no_finetuning}), where only little resources are available for criterion computation, and no fine-tuning after pruning is allowed.
Here, \gls{lrp} considerably outperformed the other metrics on toy data (cf.~Section~\ref{sec:experiments/toy_example}) and image processing benchmark data (cf.~Section~\ref{sec:no_finetuning}).
The strongly similar results between criteria observed in Scenario 1 (cf.~Section~\ref{sec:no_finetuning}) are also not surprising, as an additional file-tuning step after pruning may allow the pruned neural network model to recover its original performance, as long as the model has the capacity to do so~\cite{MolchanovTKAK16}.
}

\edited{}{
From the results of Table~\ref{tab:compareresult_vgg16_resnet50} and \editedtwo{}{Supplementary Table 3} we can observe that with a fixed pruning target of $n\%$ filters removed, \gls{lrp} might not always result in the \emph{cheapest} sub-network after pruning in terms of parameter count and \gls{flop} per inference, however it consistently is able to identify the network components for removal and preservation leading to the \emph{best performing} model after pruning.
Latter results resonate also strongly in our experiments of Scenario 2 on both image and toy data, where, without the additional fine-tuning step, the \gls{lrp}-pruned models vastly outperform their competitors.
The results obtained in multiple toy settings verify that \emph{only} the \gls{lrp}-based pruning criterion is able to preserve the original structure of the prediction function (cf.~Figures~\ref{plot:toy_data_experiment_qualitative} and~\ref{plot:toy_data_experiment_influence_of_n}).
}

\edited{}{
%Stability
Unlike the weight criterion, which is a static quantity once the network is not in training anymore, the criteria Taylor,  gradient and \gls{lrp} require reference samples for computation, which in turn may affect the estimation of neuron importance.
From the latter three criteria, however, only \gls{lrp} provides a \emph{continuous measure} of network structure importance (cf.~Sec~7.2 in~\cite{montavon2018methods}) which does not suffer from abrupt changes in the estimated importance measures with only marginal steps between reference samples.
This quality of continuity is reflected in the stability and quality of \gls{lrp} results reported in Section~\ref{sec:experiments/toy_example}, compared to the high volatility in neuron selection for pruning and model performance after pruning observable for the gradient and Taylor criteria.
From this observation it can also be deduced that \gls{lrp} requires relatively few data points to converge to a pruning solution that possesses a similar prediction behavior as the original model.
Hence, we conclude that \gls{lrp} is a robust pruning criterion that is broadly applicable in practice.
Especially in a scenario where no finetuning is applied after pruning (see Sec.~\ref{sec:no_finetuning}), the \gls{lrp} criterion allows for pruning of a large part of the model without significant accuracy drops.
}

\edited{}{
% Computational cost %marginally more than grad, en par with taylor, same order of magnitude.
In terms of computational cost, \gls{lrp} is comparable to the Taylor and Gradient criteria because these criteria require both a forward and a backward pass for all reference samples. The weight criterion is substantially cheaper to compute since it does not require to evaluate any reference samples; however, its performance falls short in most of our experiments.
Additionally, our experiments demonstrate that \gls{lrp} requires less reference samples than the other criteria (cf.~Figure~\ref{plot:toy_data_experiment_influence_of_n} \editedtwo{}{and Figure~\ref{fig:compareresult_toy}}), thus the required computational cost is lower in practical scenarios,
and better performance can be expected if only low numbers of reference samples are available (cf.~Figure~\ref{plot:result_small_sample_1}).
% lrp is (like other compared criteria) cheap to compute. given that it only requires small amounts of reference samples while still outperforming the others, the effective cost for computing pruning criteria is further decreased relatively.
}

\edited{}{
Unlike all other criteria, \gls{lrp} does not require explicit regularization via $\ell_p$-normalization,
as it is naturally normalized via its enforced \emph{relevance conservation principle} during relevance backpropagation,
which leads to the preservation of important network substructures and bottlenecks in a global model context.
In line with the findings by~\cite{MolchanovTKAK16}, our results in Figure~\ref{plot:result_vgg16} and Supplementary Figure~2 show that additional normalization after criterion computation for weight, gradient and Taylor is not only vital to obtain good performance, but also to avoid disconnected model segments --- something which is prevented out-of-the-box with \gls{lrp}.
}

%%%% Weaknesses %%%
However, our proposed criterion still provides several open questions that deserve a deeper investigation in future work.
% Dependence on implementation
First of all, \gls{lrp} is not implementation invariant, i.e., the structure and composition of the analyzed network might affect the computation of the \gls{lrp}-criterion and ``network canonization'' --- a functionally equivalent restructuring of the model --- might be required for optimal results, as discussed early in Section~\ref{Experiments} and~\cite{guillemot2020breaking}.
% LRP hyperparameters
Furthermore, while our \gls{lrp}-criterion does not require additional hyperparameters, e.g.,  for normalization, the pruning result might still depend on the chosen \gls{lrp} variant.
In this paper, we chose the $\alpha_1\beta_0$-rule in all layers,
because this particular parameterization identifies the network's neural pathways positively contributing to the selected output neurons for which reference samples are provided, is robust to the detrimental effects of shattered gradients affecting especially very deep \glspl{cnn}~\cite{montavon2017explaining} \editedtwo{}{(i.e., other than gradient-based methods, it does not suffer from potential discontinuities in the backpropagated quantities)},
and has a mathematical well-motivated foundation in \gls{dtd}~\cite{montavon2017explaining,montavon2018methods}.
However, other work from literature provide~\cite{alber2019innvestigate} or suggest~\cite{hgele2019resolving,lapuschkin2019unmasking} alternative parameterizations to optimize the method for explanatory purposes.
It is an interesting direction for future work to examine whether these findings also apply to \gls{lrp} as a pruning criterion.\\

\section{Conclusion}
\label{conclusion}
	Modern \glspl{cnn} typically have a high capacity with millions of parameters as this allows to obtain good optimization results in the training process.
	After training, however, high inference costs remain, despite the fact that the number of effective parameters in the deep model is actually significantly lower (see e.g.~\cite{murata1994network}).
	To alleviate this, pruning aims at compressing and accelerating the given models without sacrificing much  predictive performance.
	In this paper, we have proposed a novel criterion for \edited{}{the} iterative pruning of \glspl{cnn} based on the explanation method \gls{lrp}, linking for the first time two so far disconnected lines of research.
	% Clear motivation for the score, no heuristic
	\gls{lrp} has a clearly defined meaning, namely the contribution of an individual network \editedtwo{element}{unit}, i.e. weight or filter, to the network output. Removing \editedtwo{element}{unit}s according to low \gls{lrp} scores thus means discarding all aspects in the model that do not contribute relevance to its decision making.
	Hence, as a criterion, the computed relevance scores can easily and cheaply give efficient compression rates without further postprocessing, such as per-layer normalization.
	% Computational cost
    Besides, technically \gls{lrp} is scalable to general network structures and its computational cost is similar to the one of a gradient backward pass.

	% Experimental results
    In our experiments, the \gls{lrp} criterion has shown favorable compression performance on a variety of datasets both with and without retraining after pruning.
    Especially when pruning without retraining, our results for small datasets suggest that the \gls{lrp} criterion outperforms the state of the art and therefore, its application is especially recommended in transfer learning settings where only a small target dataset is available.

    % Future work
    In addition to pruning, the same method can be used to visually interpret the model and explain individual decisions as intuitive relevance heatmaps. Therefore, in future work, we propose to use these heatmaps to elucidate and explain which image features are most strongly affected by pruning to additionally avoid that the pruning process leads to undesired Clever Hans phenomena \cite{lapuschkin2019unmasking}.
    \edited{Also, we would like to further investigate \gls{lrp}-based pruning with other modern neural network architectures such as GoogLeNet, ResNet, DenseNet, etc. in order to see how to effectively prune the weights/filters in the residual dense blocks of the desired networks.
    }{}

    %pruning can be useful in the context of multi-task learning, where some hidden neurons or filters are specific to a given task while others are shared.
    %Therefore, we could renovate our models by capturing multiple components of 1) what is specifically relevant to the given task, and 2) what is commonly relevant to all tasks. This could be an interesting direction for future work.

\section*{Acknowledgements}
This work was supported by the German Ministry for Education and Research (BMBF) through BIFOLD (refs. 01IS18025A and 01IS18037A), MALT III (ref. 01IS17058), Patho234 (ref. 031L0207D) and TraMeExCo (ref. 01IS18056A), as well as the Grants 01GQ1115 and 01GQ0850;
and by Deutsche Forschungsgesellschaft (DFG) under Grant Math+, EXC 2046/1, Project ID 390685689;
 by the Institute of Information \& Communications Technology Planning \& Evaluation (IITP) grant funded by the Korea Government (No. 2019-0-00079,  Artificial Intelligence Graduate School Program, Korea University);
and by STE-SUTD Cyber Security Corporate Laboratory; the AcRF Tier2 grant MOE2016-T2-2-154;
the TL project Intent Inference;
and the SUTD internal grant Fundamentals and Theory of AI Systems.
\editedtwo{}{The authors would like to express their thanks to Christopher J Anders for insightful discussions.}

%% New version of the num-names style
\bibliographystyle{elsarticle-num-names}
\bibliography{bibliography.bib}






% \end{thebibliography}

%%%%
% SUPPLEMENARY MATERIALS BELOW
%%%%

\clearpage
\appendix
\setcounter{figure}{0}
\setcounter{table}{0}
\setcounter{page}{1}
\pagenumbering{roman}
\renewcommand{\figurename}{Supplementary Figure}
\renewcommand{\tablename}{Supplementary Table}
\renewcommand\thefigure{\arabic{figure}}
\renewcommand\thetable{\arabic{table}}

\begin{center}
	\Large
	Pruning by Explaining: A Novel Criterion for\\
	Deep Neural Network Pruning\\
	{\sc --- Supplementary Materials ---}
\end{center}

\section*{Supplementary Methods 1: Data Preprocessing}
\label{preprocessing}
During fine-tuning, images are resized to 256$\times$256 and randomly cropped to 224$\times$224 pixels, and then horizontally flipped with a random chance of 50\% for data augmentation. For testing, images are resized to 224$\times$224 pixels.

\textbf{Scene 15:} The Scene 15 dataset contains about 4\edited{}{,}485 images and consists of 15 natural scene categories obtained from COREL collection,
Google image search and personal photographs~\cite{lazebnik2006beyond}.
We fine-tuned \edited{two}{four} different models on 20\% of the images from each class %\todo[inline]{so 80\% of the data has been used for testing? is this the usual way to do this on this dataset?}
and achieved initial Top-1 accuracy of 88.59\% for VGG-16, 85.48\% for AlexNet, \edited{}{83.96\% for ResNet-18, and 88.28\% for ResNet-50,} respectively.

\textbf{Event 8:} Event-8 consists of 8 sports event categories by integrating scene and object recognition.
We use 40\% of the dataset's images for fine-tuning and the remaining 60\% for testing.
%\todo[inline]{again here: are these \emph{official} data splits?}
We adopted the common data augmentation method as in \cite{NIPS2015_5784}. %초기 성능 생략함 (어차피 테이블에 나오니까)

\textbf{Cats and Dogs:} This is the Asirra dataset provided by Microsoft Research (from Kaggle)\edited{}{.} The given dataset for the competition (KMLC-Challenge-1)~\cite{asirra2007}.
Training dataset contains 4,000 colored images of dogs and 4,005 colored images of cats, while containing 2,023 test images.
We reached initial accuracies of 99.36\% for VGG-16, 96.84\% for AlexNet, \edited{}{97.97\% for ResNet-18, and 98.42\% for ResNet-50} based on transfer learning approach.

\textbf{Oxford Flower\edited{}{s} 102:} \edited{}{The} Oxford \edited{102 flowers}{Flowers 102 dataset} contains 102 species of flower categories found in the UK, which is a collection with over 2\edited{}{,}000 training and 6\edited{}{,}100 test images~\cite{nilsback2008automated}. %The flowers appear at different scales, pose and lighting conditions.
We fine-tuned models with pre-trained networks on ImageNet for transfer learning.

\textbf{C\edited{ifar 10}{IFAR-10}:} This \edited{}{dataset} contains 50,000 training images and 10,000 test images spanning 10 categories of objects. The resolution of each image is 32$\times$32 pixels and therefore we resize the images as 224$\times$224 pixels.

\textbf{ILSVRC~2012:} In order to show the effectiveness of the pruning criteria in the small sample scenario, we pruned and tested \edited{}{all models} on randomly selected \edited{}{$k=3$ from 1000 classes and} data \edited{among}{from the ImageNet corpus~\cite{ILSVRC15}.} \edited{150,000 photographs, collected from flickr and other search engines, hand-labeled with the presence or absence of 1000 object categories~\cite{ILSVRC15}.}{}

\section*{Supplementary Results 1: Additional results on toy data}
\label{appendix-toy}

% Former Table 1 that has a new life as a figure now
\begin{table}[!ht]
    \begin{center}
        \caption{Comparison of \edited{}{training} accuracy \edited{in each criterion with pruned models on toy datasets (moon, circle, and multi-class dataset)}{after one-shot pruning one third of all filters w.r.t one of the four metrics on toy datasets, with $n \in [1, 5, 20, 100]$ reference samples used for criteria computation for \underline{W}eight, \underline{G}radient, \underline{T}aylor and \underline{L}RP. Compare to Figure 4.}}
        \label{tab:compareresult_toy}
		\setlength{\tabcolsep}{4pt}
		\renewcommand{\arraystretch}{.55}
		\scriptsize %scriptsize instead of resizebox
		%\resizebox{\linewidth}{!}{
    		\begin{tabular}{lc|c|ccc|ccc|ccc|ccc}
            \toprule
            \textbf{Dataset}    & \multicolumn{1}{c}{\textbf{Unpruned}}   & \multicolumn{13}{c}{\textbf{Pruned}} \\
            \cmidrule{3-15}
                                &                       &               & \multicolumn{3}{c|}{n=1} & \multicolumn{3}{c|}{n=5}            &   \multicolumn{3}{c|}{n=20}          &   \multicolumn{3}{c}{n=100}  \\
                                &                       &   W           &   T   &   G   &   L      &   T   &   G   &   L                 &   T   &   G   &   L                  &   T   &   G   &   L  \\
            \midrule
            moon                &           99.90       & 99.60 & 79.80 & 83.07 & \textbf{85.01}   & 84.70  & 86.07 & \textbf{99.86}     & 86.99 & 85.87 & \textbf{99.85}       & 94.77 & 93.53 & \textbf{99.85} \\
            circle              &           100.00      & 97.10 & 68.35 & 69.21 & \textbf{70.23}   & 87.18  & 82.23 & \textbf{99.89}     & 91.87 & 85.36 & \textbf{100.00}      & 97.04 & 90.88 & \textbf{100.00} \\
            multi               &           94.95       & 91.00 & 34.28 & 34.28 & \textbf{62.98}   & 77.34  & 67.96 & \textbf{91.85}     & 83.21 & 77.39 & \textbf{91.59}       & 84.76 & 82.68 & \textbf{91.25} \\
            \bottomrule
            \end{tabular}
        %} %closing bracket for \resizebox
    \end{center}
\end{table}

 % section of LRP consistency across N
    \edited{}{
    Here, we discuss with \editedtwo{}{Supplementary} Table~\ref{tab:selfrank_wrt_n_toy} the consistency of \gls{lrp}-based neuron selection across reference sample sizes.
    One can assume that the larger the choice of $n$ the less volatile is the choice of (un)important neurons by the criterion, as the influence of individual reference samples is marginalized out.
    We therefore compare the first and last ranked sets of neurons selected for a low (yet due to our observations sufficient) number $n=10$ of reference samples to all other reference sample set sizes $m$ over all unique random seed combinations.
    For all comparisons of $n \times m$ (except for $m=1$) we observe a remarkable consistency in the selection of (un)important network substructures.
    With an increasing $m$, we can see the consistency in neuron set selection gradually increase and then plateau for the ``moon'' and ``circle'' datasets, which means that the selected set of neurons remains consistent for larger sets of reference samples from that point on.
    For the ``mult'' toy dataset, we observe a gradual yet minimal decrease in the set similarity scores for $m \geq 10$, which means that the results deviate from the selected neurons for $n=10$, i.e.\ variability over the neuron sets selected for $n=10$ are the source of the volatility between $n$-reference and $m$-reference selected neuron sets.
    In all cases, peak consistency is achieved at $n\in \lbrace 5, 10\rbrace$ reference samples, identifying low numbers of $n\in \lbrace 5, 10\rbrace$ as sufficient for consistently pruning our toy models.
    }

    \begin{table}[!th]
        \begin{center}
            % CASE 4 results on github
            \caption{\edited{}{A consistency comparison of neuron selection of \gls{lrp} between reference sample sets sized $n$, averaged over all 1225 unique random seed combinations.
            Higher values indicate higher consistency. We report results for $n=10$ reference samples in comparison to $m \in [1,\editedtwo{}{2,}5,10,20,50,100,200]$ reference samples per class and $k=250$.
            Observations for all other combinations of $n \times m$, $k \in [125,500,1000]$ and all other criteria are available on \href{https://github.com/seulkiyeom/LRP_Pruning_toy_example}{github}$^3$.
            }}
            \label{tab:selfrank_wrt_n_toy}
    		\setlength{\tabcolsep}{4pt}
    		\renewcommand{\arraystretch}{.55}
    		\scriptsize %scriptsize instead of resizebox
    		\resizebox{\linewidth}{!}{
        		\begin{tabular}{ll|lcccccccc|cccccccc}
                \toprule
                \textbf{Dataset}    & $n$   &   &       \multicolumn{8}{c|}{\textbf{first-250}}                                                 & \multicolumn{8}{c}{\textbf{last-250}}\\
                                                        \cmidrule{4-11}                                                                         \cmidrule{12-19}
                                    &       & $m=$  & $1$   & \editedtwo{}{$2$}     & $5$       &   $10$    &   $20$    &   $50$    &   $100$ &   $200$                           & $1$   & \editedtwo{}{$2$}   &  $5$   &   $10$  &   $20$  &   $50$  &   $100$ &   $200$\\
                moon                & $10$  &       & 0.687 &  \editedtwo{}{0.865}  & 0.942   &   0.947   &   0.951   &   0.952   &   0.950   &   0.950                           & 0.676 & \editedtwo{}{0.810} &  0.916   &   0.928   &   0.938   &   0.946   &   0.947   &   0.948  \\
                circle              & $10$  &       & 0.689 &  \editedtwo{}{0.795}  & 0.831   &   0.843   &   0.845   &   0.846   &   0.846   &   0.842                           & 0.698 & \editedtwo{}{0.874} &  0.919   &   0.937   &   0.946   &   0.951   &   0.953   &   0.954  \\
                mult                & $10$  &       & 0.142 &  \editedtwo{}{0.625}  & 0.773   &   0.791   &   0.779   &   0.765   &   0.763   &   0.762                           & 0.160 & \editedtwo{}{0.697} & 0.890   &   0.942   &   0.940   &   0.936   &   0.934   &   0.933  \\
                \bottomrule
                \end{tabular}
            } %closing bracket for \resizebox
        \end{center}
    \end{table}

\section*{Supplementary Results 2: Additional results for image processing neural networks}
\label{appendix-images}

Here, we provide results for AlexNet and ResNet-18 --- in addition to the VGG16 and ResNet-50 results shown in the main paper --- in
Supplementary Figure~\ref{plot:perf_compare_alexnet} (cf.~\editedtwo{}{Figure~6}), Supplementary Figure~\ref{plot:result_alexnet} (cf.~\editedtwo{}{Figure~5}), and Supplementary Table~\ref{tab:compareresult_alexnet_resnet18} (cf.~\editedtwo{}{Table~3}).
These results demonstrate that the favorable pruning performance of our \gls{lrp} criterion is not limited to any specific network architecture.
\editedtwo{}{We remark that the results for CIFAR-10 show a larger robustness to higher pruning rates. This is due to the fact that CIFAR-10 has the lowest resolution as dataset and little structure in its images as a consequence. The images contain  components with predominantly very low frequencies. The filters, which are covering higher frequencies, are expected to be mostly idle for CIFAR-10. This makes the pruning task less challenging. Therefore no method can clearly distinguish itself by a different pruning strategy which addresses those filters which are covering the higher frequencies in images.
}

\begin{figure}[th!]
    	\vskip 0.2in
    	\begin{center}
    		\centerline{\includegraphics[width=0.99\linewidth]{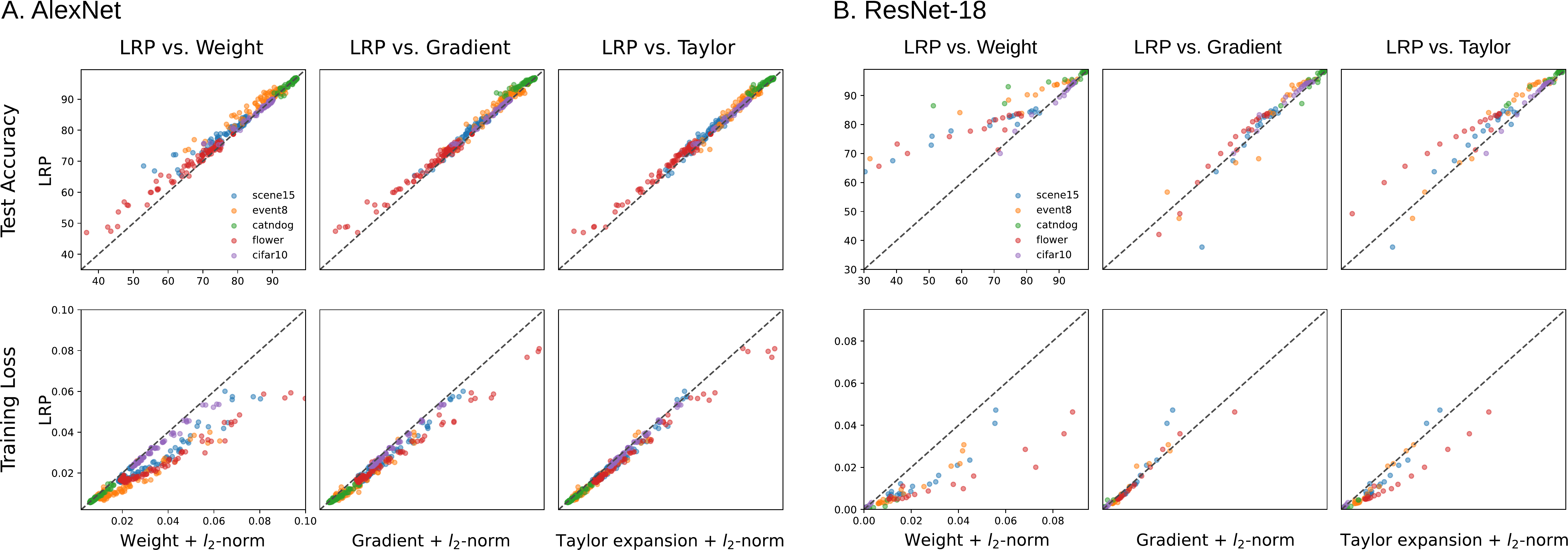}}
    		\caption{Performance comparison of the proposed method (i.e. LRP) and other criteria on AlexNet \edited{}{and ResNet-18} with five datasets.
    		\edited{}{Each point in the scatter plot corresponds to the performance at a specific pruning rate of two criteria, where the vertical axis shows the performance of LRP and the horizontal axis the performance of one other criterion (compare to Supplementary Figure~\ref{plot:result_alexnet} that displays the same data for more than two criteria). The black dashed line shows the set of points where models pruned by one of the compared criteria would exhibit identical performance to \gls{lrp}. For accuracy, higher values are better. For loss, lower values are better. Compare to \editedtwo{Figure 5}{Figure 6}.}}
    		\label{plot:perf_compare_alexnet}
    	\end{center}
    	\vskip -0.2in
    \end{figure}

    \begin{figure*}[th!]
    	\vskip 0.2in
    	\begin{center}
    		\centerline{\includegraphics[width=0.99\linewidth]{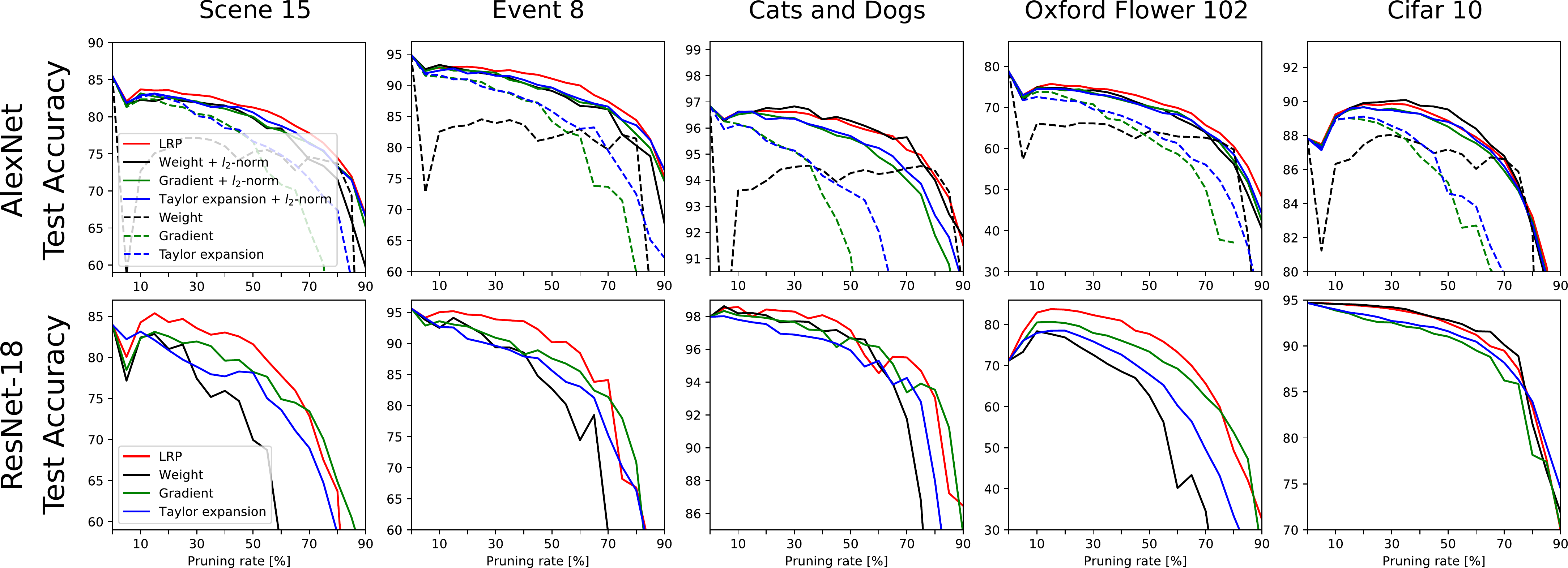}}
    		\caption{Comparison of test accuracy in different criteria as pruning rate increases on AlexNet (top) and ResNet-18 (bottom) with five datasets.
    		\edited{}{Pruning \emph{with} fine-tuning. Prematurely terminated lines indicate that during pruning, the respective criterion removed vital filters and thus disconnected the model input from the output. Compare to \editedtwo{Fig.~4.}{Figure 5.}}}
    		\label{plot:result_alexnet}
    	\end{center}
    	\vskip -0.2in
    \end{figure*}

		\begin{table}[!ht]
        \begin{center}
            \caption{
            \edited{}{
            A performance comparison between criteria (\underline{W}eight, \underline{T}aylor, \underline{G}radient with $\ell_2$-norm and \underline{L}RP) and the \underline{U}npruned model for \textbf{AlexNet} (top) and \textbf{ResNet-18} (bottom)
            on five different image benchmark datasets.
            Criteria are evaluated at fixed pruning rates per model and dataset, identified as \texttt{$\langle$dataset$\rangle$@$\langle$percent\_pruned\_filters$\rangle$\%}.
            We report test accuracy (in \%), (training) loss ($\times 10^{-2}$), number of remaining parameters ($\times 10^7)$ and FLOPs (in \gls{mmacs} for AlexNet and \gls{gmacs} for ResNet-18) per forward pass.
            For all measures except accuracy, lower outcomes are better.
            Compare to \editedtwo{Table~4}{Table~3}.
            }}
            \label{tab:compareresult_alexnet_resnet18}
    		\setlength{\tabcolsep}{4pt}
    		\renewcommand{\arraystretch}{.55}
    		\scriptsize %scriptsize AND resizebox for this one
    		\resizebox{\linewidth}{!}{
        		\begin{tabular}{l|rrrrr|rrrrr|rrrrr}
                \toprule
                \textbf{AlexNet}         &   \multicolumn{5}{c|}{\textbf{Scene 15 @ 55\%}}                                              & \multicolumn{5}{c|}{\textbf{Event 8 @ 55\%}}                                                      & \multicolumn{5}{c}{\textbf{Cats \& Dogs @ 60\%}}                                              \\
                                            \cmidrule{2-6}                                                                              \cmidrule{7-11}                                                                                     \cmidrule{12-16}
                                        &   U           &   W       &  T                &   G               &   L                       &   U           &   W       &  T                &   G               &   L                           &   U           &   W               &  T          &   G                 &   L                   \\
                Loss                    &   2.49        &   2.78    &  2.31             &   2.46            &   \textbf{2.02}           &   1.16        &   1.67    &  1.21             &   1.37            &   \textbf{1.00}               &   0.50        &   0.77            &  0.78       &   0.87              &   \textbf{0.70}       \\
                Accuracy                &   85.48       &   78.43   &  79.40            &   78.63           &   \textbf{80.76}          &   94.89       &   88.10   &  88.62            &   88.42           &   \textbf{90.41}              &   96.84       &   \textbf{95.86}  &  95.23      &   94.89             &   95.81               \\
                Params                  &   54.60       &   35.19   &  33.79            &   \textbf{33.29}  &   33.93                   &   54.57       &   34.25   &  33.00            &   33.29           &   \textbf{32.79}              &   54.54       &   32.73           &  33.50      &   33.97             &   \textbf{32.66}      \\
                FLOPs                   &   711.51      &   304.88  &  229.95           &   \textbf{225.19} &   277.98                  &   711.48      &   301.9   &  241.18           &   \textbf{238.36} &   291.95                      &   711.46      &   264.04          &  199.88     &   \textbf{190.84}   &   240.19              \\
                \multicolumn{1}{c}{}    & \multicolumn{5}{c}{}                                                                      & \multicolumn{5}{c}{}                                                                              & \multicolumn{5}{c}{}                                                                          \\
                                        & \multicolumn{5}{c|}{\textbf{Oxford Flower 102 @ 70\%}}                                    &   \multicolumn{5}{c|}{\textbf{CIFAR-10 @ 30\%}}                                           & \multicolumn{5}{c}{}\\
                                        \cmidrule{2-6}                                                                                  \cmidrule{7-11}
                                        &   U       &   W               &  T                &   G       &   L                       &   U       &   W               &  T                &   G               &   L               &      &      &      &      &      \\
                Loss                    &    5.15   &    4.83           &   3.39            &   3.77    &   \textbf{3.01}           &   1.95    &   2.44            &  2.46             &   2.46            &   \textbf{2.33}   &      &      &      &      &      \\
                Accuracy                &   78.74   &   63.93           &  64.10            &   64.11   &   \textbf{65.69}          &   87.83   &   \textbf{90.03}  &  89.48            &   89.58           &   89.87           &      &      &      &      &      \\
                Params                  &   54.95   &   \textbf{28.13}  &  29.19            &   28.72   &   28.91                   &   54.58   &   48.31           &  84.23            &   \textbf{46.31}  &   48.22           &      &      &      &      &      \\
                FLOPs                   &   711.87  &   192.69          &  \textbf{132.34}  &  141.82   &   161.35                  &   711.49  &   477.16          &  \textbf{371.48}  &   395.43          &   402.93          &      &      &      &      &      \\
                \bottomrule
                % ABOVE: AlexNet. BELOW: ResNet-18
                \toprule
                \textbf{ResNet-18}         &   \multicolumn{5}{c|}{\textbf{Scene 15 @ 50\%}}                                           & \multicolumn{5}{c|}{\textbf{Event 8 @ 55\%}}                                                      & \multicolumn{5}{c}{\textbf{Cats \& Dogs @ 45\%}}                                  \\
                                            \cmidrule{2-6}                                                                            \cmidrule{7-11}                                                                                     \cmidrule{12-16}
                                        &   U           &   W       &  T                &   G           &   L                       &   U           &   W       &  T                &   G               &   L                           &   U           &   W       &  T          &   G             &   L                   \\
                Loss                    &   1.32           &   1.98       &  1.28                &   1.03           &   \textbf{0.85}                       &     0.61        &   1.28       &  0.99                &  0.72               &   \textbf{0.55}                           &   0.03           &   0.04       &  0.05          &   0.04             &   \textbf{0.02}                   \\
                Accuracy                &   83.97           &   69.95       &  78.14                &   78.24           &   \textbf{81.61}                       &   95.63           &         80.20       &  83.81                &   86.76               &   \textbf{90.27}                            &   97.97           &   97.17       &  96.34          &   94.13             &   \textbf{97.91}                   \\
                Params                  &   11.18           &   4.63       &  4.91                &   4.96           &   \textbf{4.52}                       &   11.18           &   3.99       &  4.17                &   4.26               &   \textbf{3.89}                           &   11.18           &   \textbf{4.88}       &  5.18          &   5.15             &   5.04                   \\
                FLOPs                   &   1.82           &   1.30       &  1.16                &   \textbf{1.10}           &   1.27                       &   1.82           &   1.22       &  1.11                &   \textbf{1.07}               &   1.20                           &   1.82           &   1.36       &  1.22          &   \textbf{1.21}             &   1.36                   \\
                \multicolumn{1}{c}{}    & \multicolumn{5}{c}{}                                                                      & \multicolumn{5}{c}{}                                                                              & \multicolumn{5}{c}{}                                                          \\
                                        & \multicolumn{5}{c|}{\textbf{Oxford Flower 102 @ 70\%}}                        &   \multicolumn{5}{c|}{\textbf{CIFAR-10 @ 30\%}}     & \multicolumn{5}{c}{}\\
                                        \cmidrule{2-6}                                                                      \cmidrule{7-11}
                                        &   U       &   W       &  T            &   G       &   L                       &   U       &   W           &  T            &   G       &   L               &      &      &      &      &      \\
                Loss                    &   1.36       &   4.64       &  2.96            &   1.65       &   \textbf{1.59}                       &   0.000       &   \textbf{0.002}           &  0.012            &   0.016       &   0.004               &      &      &      &      &      \\
                Accuracy                &   71.23       &   34.58       &  49.56            &   62.41       &   \textbf{65.60}                       &   94.67       &   \textbf{94.21}           &  92.71            &   92.55       &   94.03               &      &      &      &      &      \\
                Params                  &   11.23       &   \textbf{2.19}       &  3.07            &   3.07       &   2.45                       &   11.17       &   6.88           &  7.85            &   7.56       &   \textbf{6.62}               &      &      &      &      &      \\
                FLOPs                   &   1.82       &   0.95       &  \textbf{0.72}            &   0.73       &   0.93                       &   0.56       &   0.46           &  0.42            &   \textbf{0.41}       &   0.47               &      &      &      &      &      \\
                \bottomrule
                \end{tabular}
            } %closing bracket for \resizebox
        \end{center}
    \end{table}

    \editedtwo{}{
	Furthermore, the ASSIRA dataset of cats and dogs may raise the concern that it is the MNIST analogue for pets, representing a rather simple problem, and for that reason the results might not be transferable to problems with larger variance.
	To validate our observation,
	we have chosen the more lightweight~\cite{BiancoModelComparison} ResNet-50 model,
	and evaluated its pruning performance on three further datasets.
	Each of the three datasets is composed as a binary discrimination problem obtained by fusing two datasets chosen from the following selection:
	FGVC aircraft~\cite{maji13fine-grained}, CUB birds~\cite{WahCUB_200_2011} and Stanford cars \cite{KrauseStarkDengFei-Fei_3DRR2013}.
	We have chosen these three datasets,
	as they are known from the literature,
	have an intrinsic variability as visible by their numbers of classes and a medium sample size.
	Most importantly,
	we know for these datasets that the object categories defining their contents are matched by some classes in the ImageNet dataset which is used to initialize the weights of the ResNet-50 network.
	}

	\editedtwo{}{
	Supplementary Figure \ref{plot:result_resnet} shows the results for the three composed discrimination problems, without fine-tuning after pruning. We can observe that each pruning method is able to remove a certain amount of network filters without notable loss of discrimination performance. Weight-based pruning performs second best, while LRP-based pruning allows consistently to prune the largest fraction of filters before starting to lose prediction accuracy.
	    \begin{figure}[t!]
		\begin{center}
			\centerline{\includegraphics[width=\linewidth]{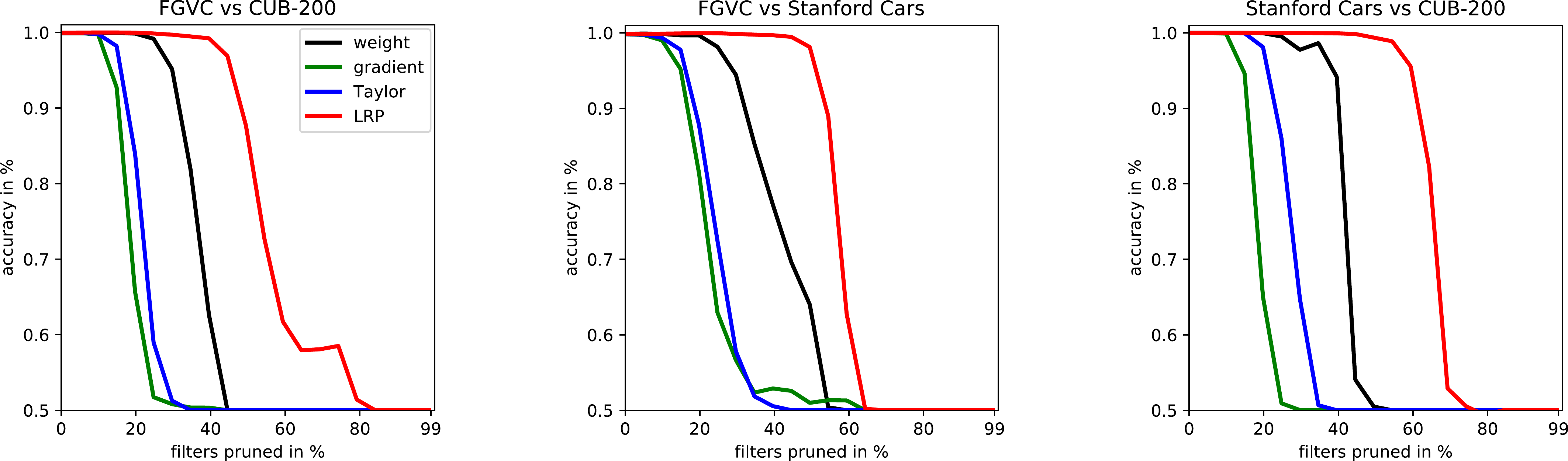}}
			\caption{
			\editedtwo{}{
			Comparison of pruning performance without a subsequent finetuning step for a ResNet-50 network when pruned by criteria using weights, gradient, Taylor-expansion and LRP. Each dataset is a binary classification problem created by combining two datasets taken from FGVC Aircraft \cite{maji13fine-grained}, CUB-200-2011 birds \cite{WahCUB_200_2011} and Stanford Cars \cite{KrauseStarkDengFei-Fei_3DRR2013}, which are covered by similar classes in the ImageNet initialization. Results are the average of 20 repetitions with randomly drawn samples. Each run relies on 20 samples for pruning, 10 from each of the two datasets, and 2048 samples for test accuracy evaluation. For a given repetition, all methods use the same set of samples for pruning and they use a set of samples for evaluation, which is again identical for all pruning methods, but disjoint from the pruning set. Compare to Figure 10.}
			}
			\label{plot:result_resnet}
		\end{center}
	\end{figure}
	}

	\editedtwo{}{
	When comparing cats versus dogs in Figure~10 against the three composed datasets in Supplementary Figure~\ref{plot:result_resnet}, we observe that there is less redundant capacity which can be pruned away for the composed datasets. This sanity check is in line with the higher variance of these composed datasets as compared to cats versus dogs. As a side remark, this observation suggests the thought to measure empirically dataset complexities with respect to a neural network by the area under the accuracy graph with respect to the amount of pruned filters.
    }

\end{document}